\documentclass[11pt, letterpaper]{paper}

\usepackage[all]{hypcap}
\usepackage[comma,numbers,sort,compress]{natbib}
\usepackage{hyperref}

\hypersetup{
    colorlinks = true,
    citecolor = {digitalred},
    linkcolor = {deepblue},
}

\usepackage{microtype}
\usepackage{graphicx}
\usepackage{booktabs}
\usepackage{float}
\usepackage{tabularx}
\usepackage{amsmath}
\usepackage{amssymb}
\usepackage{mathtools}
\usepackage{amsthm}
\usepackage{mathrsfs}
\usepackage{nicefrac}
\usepackage{dsfont}
\usepackage{enumitem}
\usepackage{float}

\usepackage{xspace}
\usepackage[capitalize,noabbrev]{cleveref}
\usepackage{subcaption}
\usepackage{wrapfig}
\usepackage{listings}

\usepackage{amsmath}
\usepackage{amssymb}
\usepackage{mathtools}
\usepackage{amsthm}
\usepackage{bbm}
\usepackage{tabularx} 
\usepackage{multirow}
\usepackage[export]{adjustbox}
\usepackage{colortbl}

\usepackage{algpseudocode}
\usepackage{setspace}

\usepackage{color}
\definecolor{deepblue}{rgb}{0,0,0.5}
\definecolor{deepred}{rgb}{0.6,0,0}
\definecolor{deepgreen}{rgb}{0,0.5,0}

\definecolor{MyBlue}{RGB}{32,102,214}
\definecolor{MyRed}{RGB}{200,60,50}

\usepackage[capitalize,noabbrev]{cleveref}

\definecolor{lightblue}{rgb}{0.22,0.45,0.70}

\usepackage[most]{tcolorbox}

\newtcolorbox{tipbox}{
  enhanced,
  boxrule=0.6pt,
  left=6pt,right=6pt,top=6pt,bottom=6pt,
  colback=blue!3,
  colframe=blue!40!black,
  arc=3pt,
}

\newcommand\pythonstyle{\lstset{
basicstyle=\ttfamily\footnotesize,
language=Python,
morekeywords={self, clip, exp, mse_loss, uniform_sample, concatenate, logsumexp},
keywordstyle=\color{deepblue},
emph={MyClass,__init__},
emphstyle=\color{deepblue},
stringstyle=\color{deepgreen},
frame=single,
showstringspaces=false
}}

\lstnewenvironment{python}[1][]
{
\pythonstyle
\lstset{#1}
}
{}

\newcommand\pythoninline[1]{{\pythonstyle\lstinline!#1!}}

\definecolor{blanchedalmond}{rgb}{1.0, 0.92, 0.8}
\definecolor{carmine}{rgb}{0.59, 0.0, 0.09}
\definecolor{lightblue}{rgb}{0.22,0.45,0.70}

\renewcommand{\mathbf}{\boldsymbol}

\makeatletter
\def\Ddots{\mathinner{\mkern1mu\raise\p@
\vbox{\kern7\p@\hbox{.}}\mkern2mu
\raise4\p@\hbox{.}\mkern2mu\raise7\p@\hbox{.}\mkern1mu}}
\makeatother

\numberwithin{equation}{section}

\definecolor{amaranth}{rgb}{0.9, 0.17, 0.31}
\definecolor{antiquebrass}{rgb}{0.8, 0.58, 0.46}
\definecolor{antiquefuchsia}{rgb}{0.57, 0.36, 0.51}
\definecolor{chromeyellow}{rgb}{0.31, 0.47, 0.26}

\usepackage[dvipsnames]{xcolor}
\definecolor{maj5}{HTML}{2b8cbe}
\definecolor{maj5Imp}{HTML}{084081}

\definecolor{seq5wo}{HTML}{d95f0e}
\definecolor{seq5woImp}{HTML}{662506}

\definecolor{seq5w}{HTML}{6a51a3}
\definecolor{seq5wImp}{HTML}{3f007d}

\definecolor{selfwo}{HTML}{d95f0e}
\definecolor{selfwoImp}{HTML}{662506}

\definecolor{selfw}{HTML}{6a51a3}
\definecolor{selfwImp}{HTML}{3f007d}

\definecolor{glorewo}{HTML}{d95f0e}
\definecolor{glorewoImp}{HTML}{662506}

\definecolor{glorew}{HTML}{6a51a3}
\definecolor{glorewImp}{HTML}{3f007d}

\definecolor{vstar}{HTML}{d95f0e}
\definecolor{vstarImp}{HTML}{662506}

\makeatletter
\def\mathcolor#1#{\@mathcolor{#1}}
\def\@mathcolor#1#2#3{%
  \protect\leavevmode
  \begingroup
    \color#1{#2}#3%
  \endgroup
}
\makeatother

\usepackage{algorithm}
\usepackage{amssymb}

\Crefformat{equation}{#2Eq.\;(#1)#3}
\Crefformat{figure}{#2Figure #1#3}

\usepackage{crossreftools}
\usepackage{pifont}
\usepackage{soul}
\newcommand{\cmark}{\ding{51}}
\newcommand{\xmark}{\ding{55}}

\newif\ifcomments
\commentsfalse
\ifcomments
    \providecommand{\ion}[1]{{\color{blue}{/* ion: #1 */}}}
\else
    \providecommand{\ion}[1]{}
\fi

\definecolor{BerkeleyBlue}{HTML}{9a1f36}

\definecolor{digitalred}{rgb}{0.694,0.016,0.055}
\renewcommand{\titlefont}{\color{digitalred}\normalfont\bfseries\fontsize{16.6}{19.1}\selectfont}

\definecolor{linkgray}{HTML}{6E6F5E}

\definecolor{darknavy}{HTML}{00134D}

\renewcommand{\absfont}{\linespread{1.1}\fontsize{10.8}{13.3}\color{black}\selectfont}

\title{LLM-as-a-Verifier: A General-Purpose Verification Framework}

\makeatletter
\let\@author\@empty
\renewcommand{\author}[1]{%
  \ifx\@author\@empty
    \gdef\@author{\Authfont #1}%
  \else
    \expandafter\gdef\expandafter\@author\expandafter{\@author\\[0.4em]\Authfont #1}%
  \fi
}
\makeatother

\author{
\textbf{Jacky Kwok}$^{1}$, \,
\textbf{Shulu Li}$^{2}$, \,
\textbf{Pranav Atreya}$^{2}$, \,
\textbf{Yuejiang Liu}$^{1}$, \,
\textbf{Yixing Jiang}$^{1}$ \\[4pt]
\textbf{Chelsea Finn}$^{1}$, \,
\textbf{Marco Pavone}$^{1,3}$, \,
\textbf{Ion Stoica}$^{2}$, \,
\textbf{Azalia Mirhoseini}$^{1}$ \\
\vspace{0.4em}
{\Affilfont
$^{1}$Stanford University \,\,
$^{2}$UC Berkeley \,\,
$^{3}$NVIDIA Research}
}
\date{}

\correspondingauthor{Jacky Kwok \texttt{<jackykwok@stanford.edu>}}

\makeatletter
\fancypagestyle{firststyle}{%
  \fancyhf{}%
  \fancyfoot[L]{\raisebox{8pt}[0pt][0pt]{\parbox[t]{\textwidth}{%
    {\footerfont\fontsize{10.8}{12.8}\selectfont\upshape\bfseries
     \begingroup\hypersetup{urlcolor=darknavy}%
     \href{https://llm-as-a-verifier.com}{\raisebox{-0.25em}{\includegraphics[height=1em]{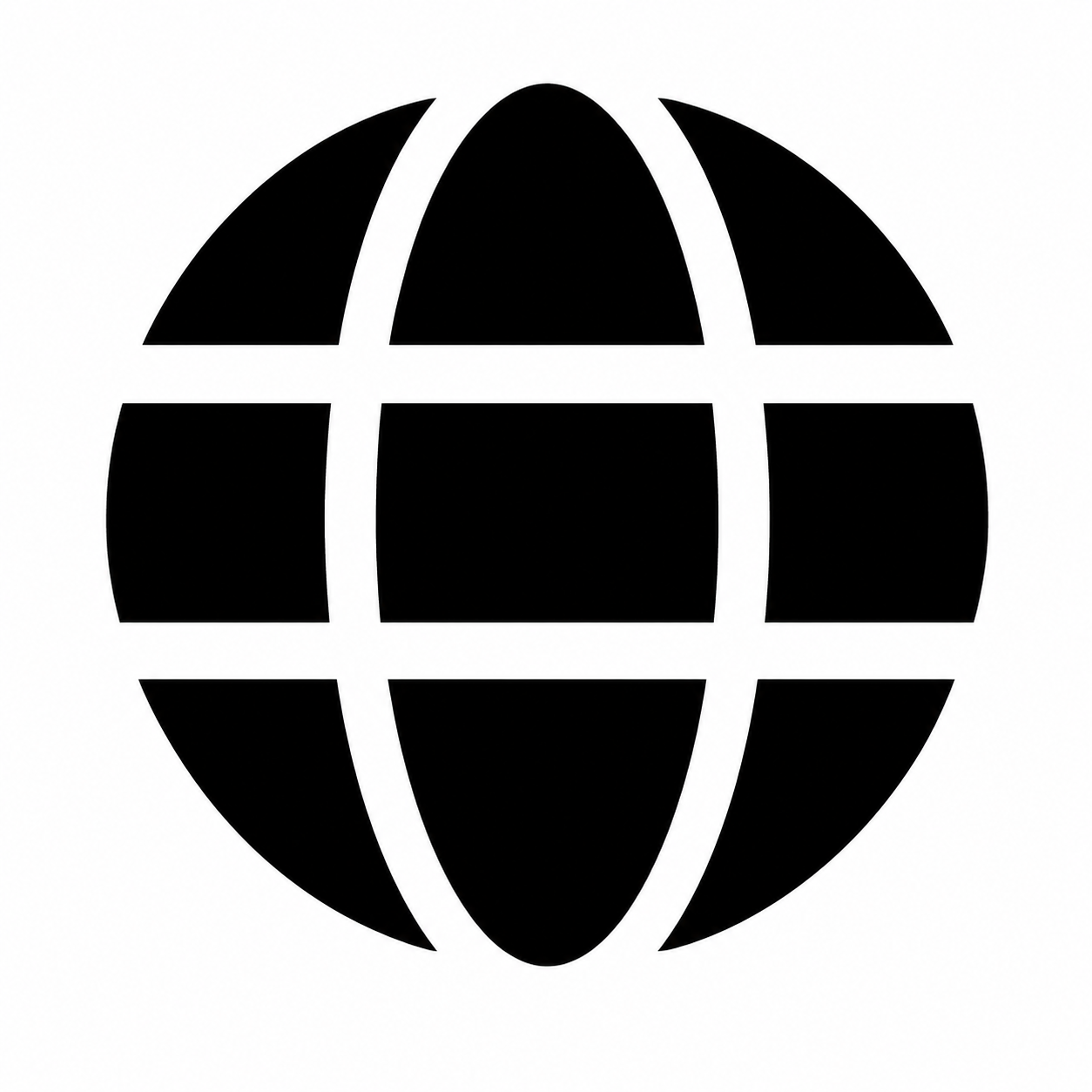}}~\textcolor{darknavy}{\texttt{\bfseries llm-as-a-verifier.com}}}\quad
     \href{https://github.com/llm-as-a-verifier/llm-as-a-verifier}{\raisebox{-0.25em}{\includegraphics[height=1em]{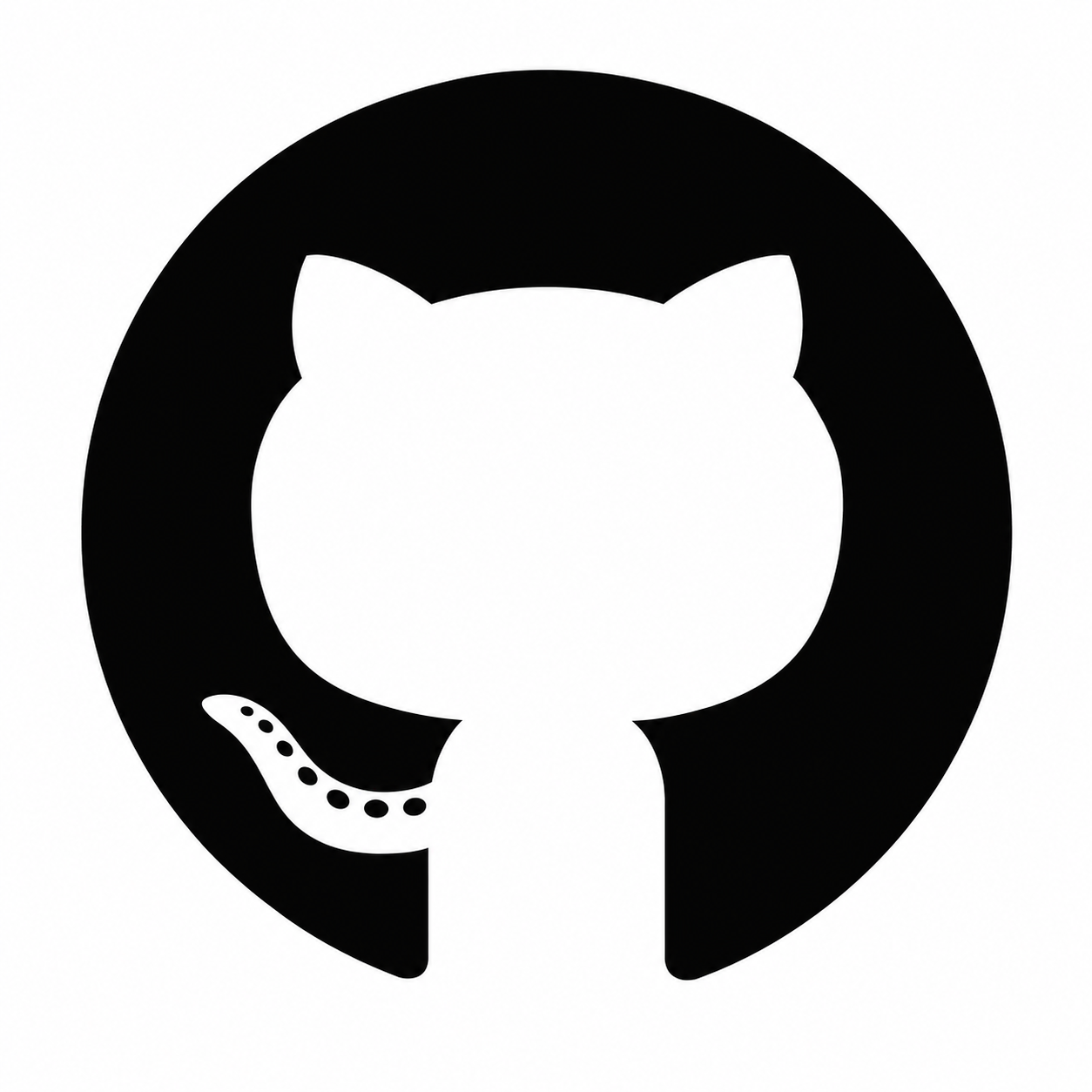}}~\textcolor{darknavy}{\texttt{\bfseries llm-as-a-verifier}}}\quad
     \href{https://github.com/llm-as-a-verifier/TurboAgent}{\raisebox{-0.25em}{\includegraphics[height=1em]{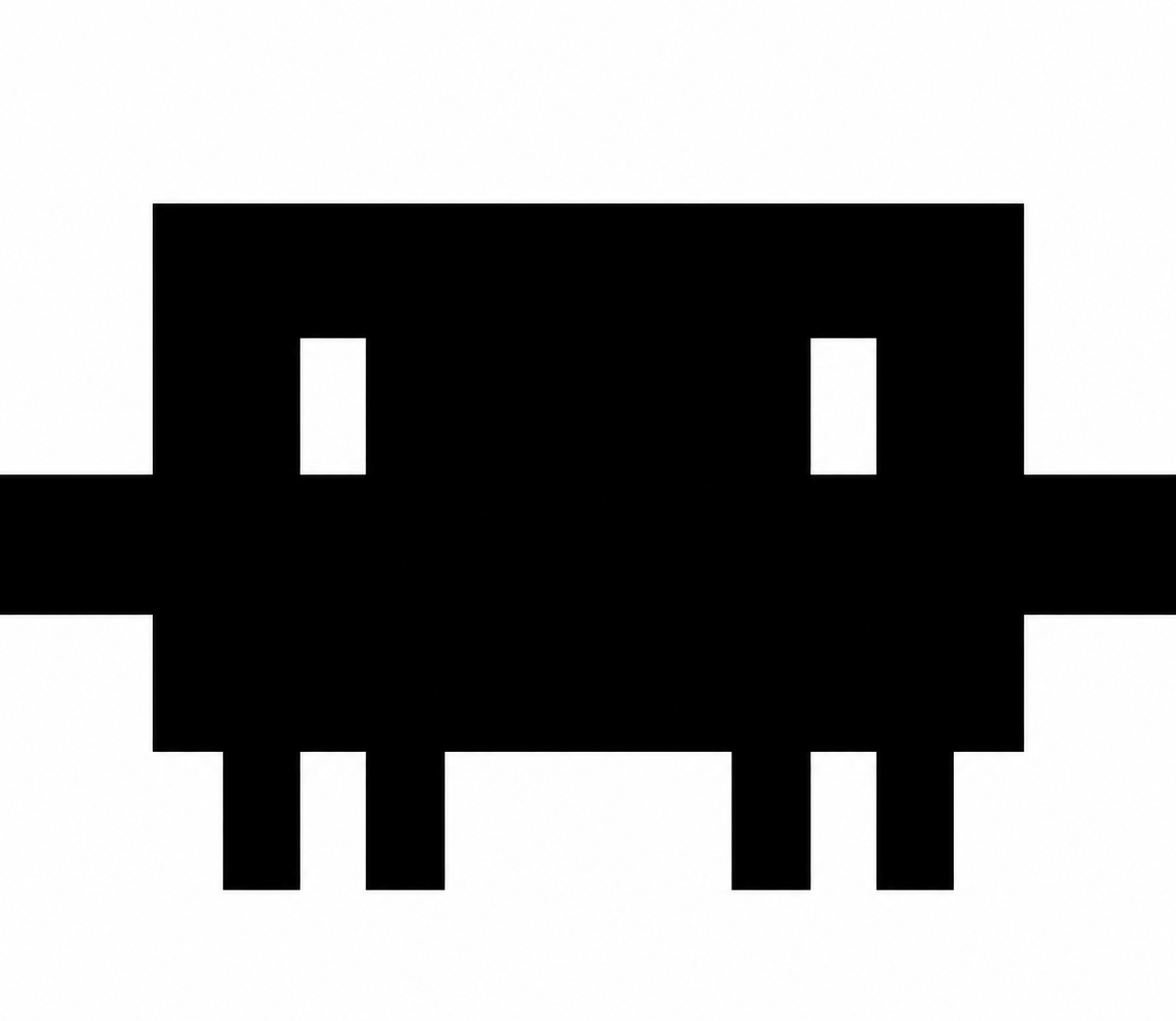}}~\textcolor{darknavy}{\texttt{\bfseries claude-code-extension}}}%
     \endgroup}\\[3pt]
    \ifdefined\correspondingauthor
      \if\relax\the\correspondingauthor\relax\else
        {\footerfont\itshape Corresponding author: \the\correspondingauthor}%
      \fi
    \fi
  }}}%
}
\makeatother

\begin{abstract}
\end{abstract}

\begin{document}

\maketitle

\vspace{-20pt}

\begin{figure}[h]
    \centering    \includegraphics[width=\linewidth]{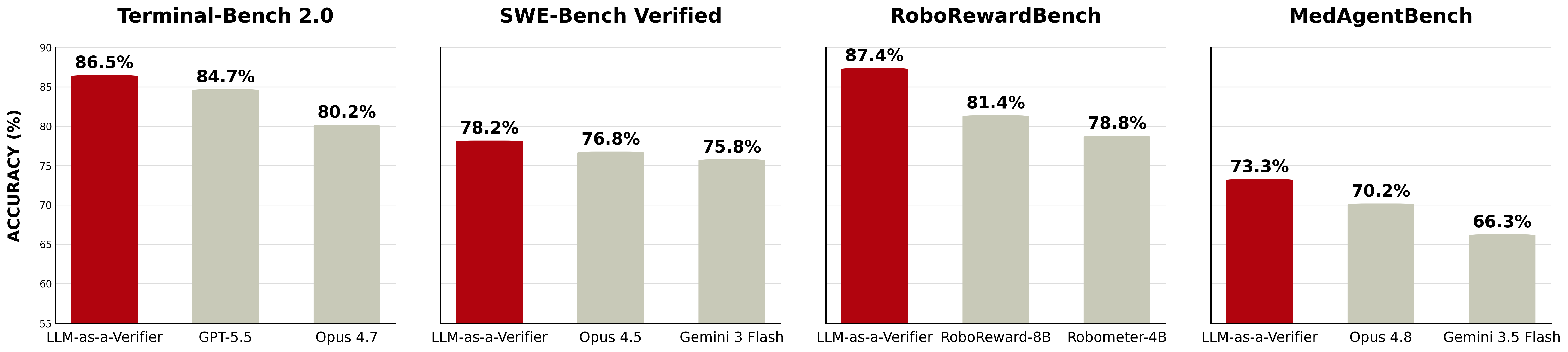}
    \caption{\textbf{Overall Performance Results.} Our proposed framework, \textbf{LLM-as-a-Verifier}, achieves state-of-the-art performance across coding, robotics, and medical domains: Terminal-Bench V2 (86.5\%), SWE-Bench Verified (78.2\%), RoboRewardBench (87.4\%), and MedAgentBench (73.3\%).}
    \label{fig:teaser}
\end{figure}

\vspace{4pt}

{\absfont \noindent\textbf{Abstract:} Scaling pre-training, post-training, and test-time compute have become the central paradigms for improving the capabilities of large language models (LLMs). In this work, we identify verification—the ability to determine the correctness of a solution—as a new scaling axis. To unlock this and demonstrate its effectiveness, we introduce LLM-as-a-Verifier, a general-purpose verification framework that provides fine-grained feedback for agentic tasks without requiring additional training. Unlike standard LM judges that prompt LLMs to produce discrete scores for candidate solutions, LLM-as-a-Verifier computes the expectation over the distribution of scoring token logits to generate continuous scores. This probabilistic formulation substantially reduces tie rates when comparing complex solutions and enables verification to scale along multiple dimensions: (1) score granularity, (2) repeated evaluation, and (3) criteria decomposition. In particular, we show that scaling the scoring granularity leads to better separation between positive and negative solutions, resulting in more calibrated comparisons. Moreover, scaling repeated evaluation and criteria decomposition consistently leads to additional gains in verification accuracy through variance and complexity reduction. To make verification scaling practical, we further introduce a cost-efficient ranking algorithm for selecting the best solution among candidates using the preference probabilities derived from the verifier's continuous scores. LLM-as-a-Verifier is effective across coding, robotics, and medical domains. It achieves state-of-the-art performance on Terminal-Bench V2 (86.5\%), SWE-Bench Verified (78.2\%), RoboRewardBench (87.4\%), and MedAgentBench (73.3\%). Beyond verification, the fine-grained signals from LLM-as-a-Verifier can also serve as a proxy for estimating task progress. We build extensions for Claude Code and Codex, enabling developers to monitor and improve their own agentic systems. Finally, we show that LLM-as-a-Verifier can be used as a dense reward signal for RL, improving the sample efficiency of SAC and GRPO on robotics and mathematical reasoning benchmarks. \par} 
\clearpage

\begin{figure}[t!]
    \centering
    \includegraphics[width=\linewidth]{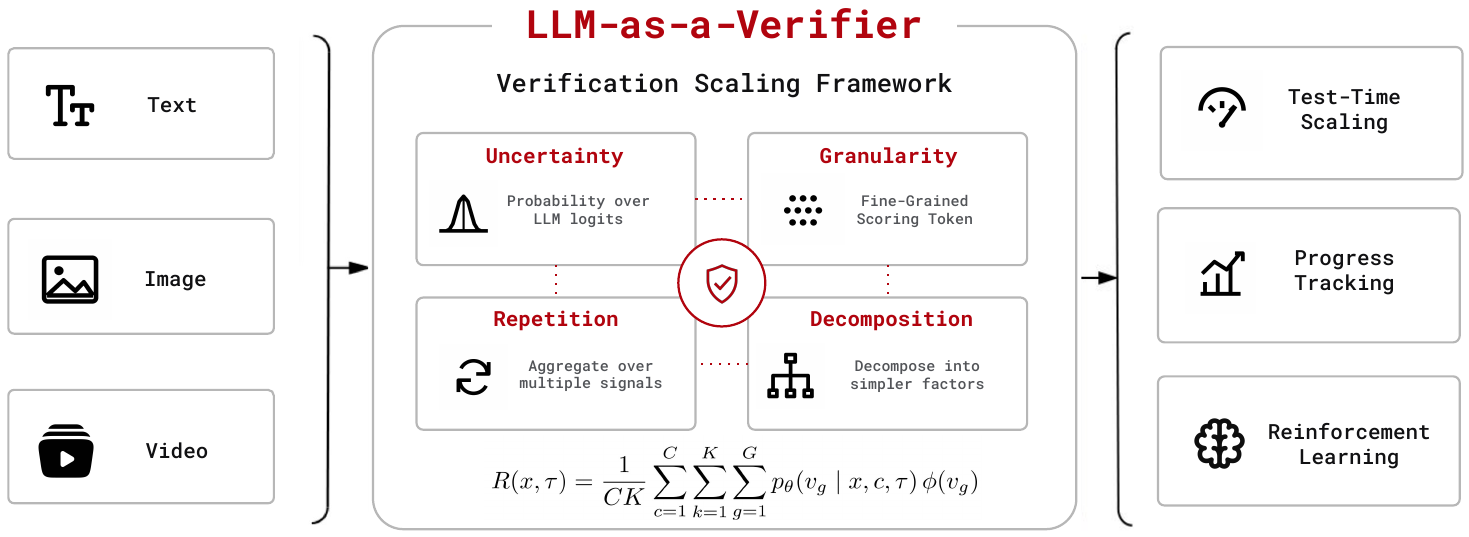}
    \caption{\textbf{Multiple modalities, many applications, one unified verification framework.} We present LLM-as-a-Verifier, a general-purpose framework that provides fine-grained feedback for any modality without requiring additional training. By leveraging the full distribution of scoring-token logits, our method captures evaluation uncertainty and enables verification to scale along three dimensions: score granularity, repeated evaluation, and criteria decomposition. The resulting fine-grained feedback can be used for test-time scaling, progress tracking, and reinforcement learning.}
    \label{fig:overview}
\end{figure}

\section{Introduction}
Recent advances in large language models (LLMs) have established scaling as a central paradigm for improving their capabilities. Performance has been driven by scaling along multiple axes, including pre-training data and compute, post-training optimization, and test-time inference~\citep{kaplan_scaling_2020, gao_scaling_2022, snell_scaling_2024}. \begin{wrapfigure}{r}{0.283\textwidth}
    \centering
    \vspace{2pt}
    \includegraphics[width=\linewidth]{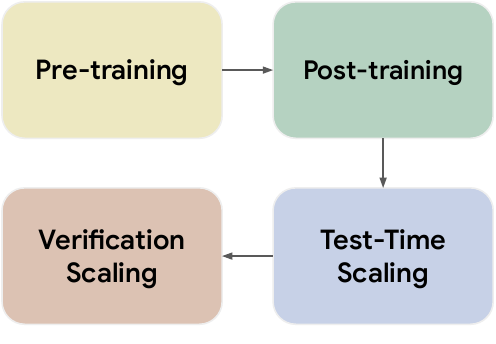}
    \caption{Scaling paradigms for large language models.}
    \label{fig:paradigms}
\end{wrapfigure}However, while generation has benefited significantly from these scaling paradigms, verification---the ability to determine the quality or correctness of a solution---has not seen the same degree of scaling. In this work, we argue that verification itself constitutes a distinct and underexplored scaling axis. Unlike generation, which benefits from well-established scaling laws, verification in current systems remains fundamentally limited. In particular, standard LM judges collapse scoring distributions into coarse discrete scores~\citep{zheng_judging_nodate, singh_v_1_2026}, leading to ties and poor discrimination, while learned reward models are constrained by training data and often fail to generalize across domains~\citep{zhang2025generativeverifiersrewardmodeling, cobbe_training_2021}. These limitations hinder the scalability of verification, preventing further performance improvements.

\begin{figure}[t!]
    \centering
    \includegraphics[width=\linewidth]{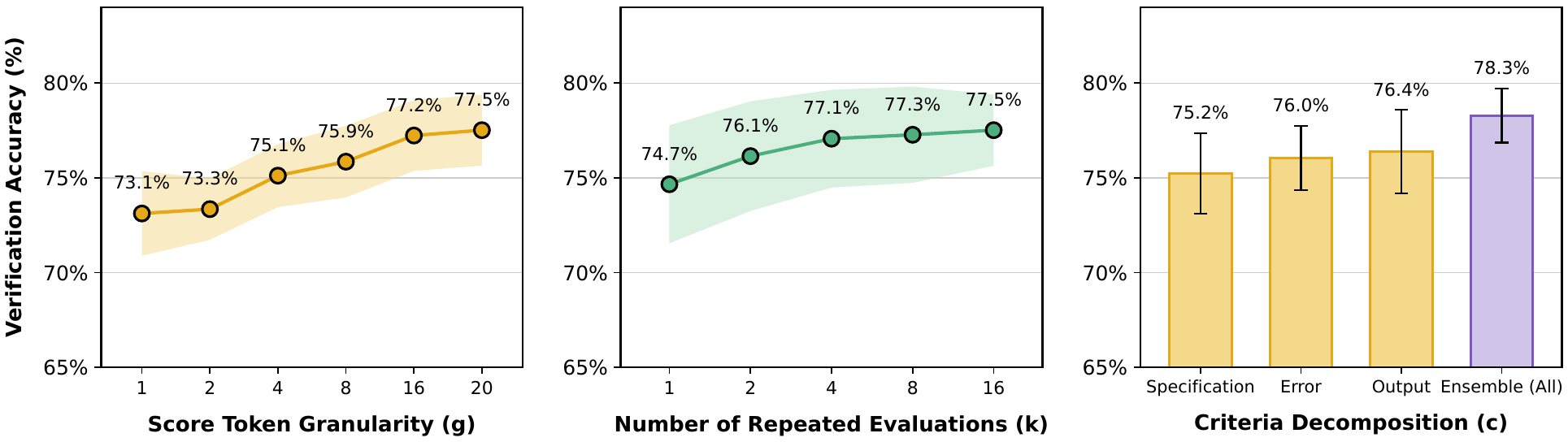}
    \caption{\textbf{Verification Scaling.} We find that verification accuracy consistently improves as we scale across multiple dimensions: (1) the granularity of score tokens, (2) the number of repeated evaluations, and (3) the decomposition of evaluation criteria. Verification accuracy is measured as the pairwise accuracy of the verifier in assigning a higher score to the ground-truth successful solution than to failed solutions for the same task on Terminal-Bench V2.}
    \label{fig:scaling_verifier}
\end{figure}

To this end, we introduce LLM-as-a-Verifier, a general-purpose verification framework that provides dense and fine-grained feedback without requiring additional training. Unlike traditional approaches that prompt LLMs to produce discrete scores within the language space~\citep{zheng_judging_nodate}, LLM-as-a-Verifier estimates the quality of candidate solutions by computing the expectation over the distribution of scoring token logits. In Fig.~\ref{fig:scaling_verifier}, we show that this probabilistic formulation unlocks multiple axes of scaling for verification. We first demonstrate that scaling the number of extracted token logits consistently reduces the tie rate when comparing complex solutions and improves the separation between positive and negative solutions. We observe that an individual evaluation or a single criterion can be biased or noisy. To mitigate this, we scale verification along two additional dimensions, repeated evaluations (which reduces variance) and criteria decomposition (which reduces prompt bias), leading to higher verification accuracy. We quantify these scaling benefits under controlled budgets, comparing LLM-as-a-Verifier against a discrete LM judge baseline in Sec.~\ref{sec:scaling}. To make verification scaling practical, we further introduce a cost-efficient ranking algorithm for selecting the best solution among candidates using the preference probabilities derived from the verifier’s continuous scores.

Interestingly, we find that the fine-grained signals produced by LLM-as-a-Verifier enable the evaluation of entire interaction trajectories rather than only intermediate steps or final outcomes as in PRMs and ORMs~\citep{cobbe_training_2021, lightman_lets_2023} for agentic tasks. When used as a trajectory reward model with our cost-efficient ranking algorithm, LLM-as-a-Verifier outperforms frontier models on challenging benchmarks across coding, robotics, and medical domains. It achieves state-of-the-art performance on Terminal-Bench V2 (86.5\%), SWE-Bench Verified (78.2\%), RoboRewardBench (87.4\% Trajectory Preference Accuracy), and MedAgentBench (73.3\%). 

Beyond its role as a verifier, our approach can also serve as a proxy for estimating task progress. Notably, we observe a strong correlation between the chronological order of steps and the verifier score (Fig.~\ref{fig:voc-counterfactual-code}). To instantiate these capabilities, we provide extensions for Claude Code and Codex, enabling users to monitor task progress and harness the benefits of LLM-as-a-Verifier to improve their own agentic systems. In robotics, our approach outperforms state-of-the-art reward models, including Robometer~\citep{liang2026robometer}, TOPReward~\citep{chen2026topreward}, and RoboReward~\citep{lee2026roboreward}, achieving a mean Value-Order Correlation (VOC) of 0.966. Overall, LLM-as-a-Verifier provides a scalable mechanism for improving the evaluation and monitoring of autonomous agents and robots in real-world environments.

Additionally, we demonstrate that using LLM-as-a-Verifier as a dense reward signal improves the sample efficiency of both off-policy and on-policy reinforcement learning algorithms. On LIBERO~\cite{liu2023liberobenchmarkingknowledgetransfer}, LLM-as-a-Verifier achieves $\approx1.8\times$ higher sample efficiency than sparse reward baselines when fine-tuning a $\pi_0$ policy with DSRL-SAC~\cite{wagenmaker2025steeringdiffusionpolicylatent}, while also reaching a higher final success rate. On the MATH reasoning benchmark, it achieves $\approx1.1\times$ higher sample efficiency when fine-tuning Qwen3-8B with GRPO~\cite{deepseek-ai_deepseek-r1_2025}.
\clearpage

In summary, our contributions are as follows:

\begin{enumerate}
\item We introduce LLM-as-a-Verifier, a probabilistic verification framework that leverages the full distribution of scoring token logits to produce fine-grained feedback and characterize three key axes of verification scaling: (1) score granularity, (2) repeated evaluation, and (3) criteria decomposition.

\item 
We propose a cost-efficient algorithm for ranking candidates and demonstrate that, when combined with verification scaling, LLM-as-a-Verifier achieves state-of-the-art performance across coding, robotics, and medical benchmarks without requiring additional training.

\item
We show that the fine-grained verifier score correlates with an agent's task progress and can be used to monitor the behavior of agents and robots.

\item
We demonstrate that LLM-as-a-Verifier can provide dense feedback for reinforcement learning, improving the sample efficiency of both on-policy and off-policy algorithms across robotics and mathematical reasoning benchmarks.

\end{enumerate}
\section{Preliminaries}
We model an agent interacting with an environment as a finite-horizon Markov Decision Process (MDP) $\mathcal{M} = (\mathcal{C}, \mathcal{S}, \mathcal{A}, P, R, H)$, where $\mathcal{C}$ denotes the space of contexts, $\mathcal{S}$ the state space, $\mathcal{A}$ the action space, $P : \mathcal{C} \times \mathcal{S} \times \mathcal{A} \rightarrow \Delta(\mathcal{S})$ the transition dynamics, $R : \mathcal{C} \times \mathcal{S} \times \mathcal{A} \rightarrow \mathbb{R}$ the reward function, and $H \in \mathbb{N}^+$ the horizon. At the beginning of each episode, a task prompt $x \in \mathcal{C}$ is sampled, and the agent begins in an initial state $s_1 \in \mathcal{S}$. At each timestep $t \in [1, H]$, the agent observes the current state $s_t$, selects an action $a_t \in \mathcal{A}$, and transitions to the next state $s_{t+1} \sim P(\cdot \mid x, s_t, a_t)$. In LLM-based agents, states correspond to prior interaction histories, and actions correspond to token sequences, such as natural language responses, code edits, and tool calls. A trajectory is defined as $\tau = (s_1, a_1, s_2, a_2, \dots, s_H, a_H)$. We assume access to a language model $\pi_\theta : \mathcal{C} \times \mathcal{S} \rightarrow \Delta(\mathcal{A})$, parameterized by $\theta$, from which actions are sampled autoregressively. A reward model assigns a scalar score to actions or trajectories. Conventional approaches rely on prompting LLMs to produce discrete scores in the language space. Formally, such reward models can be written as $R_{\text{LM}}(x, \tau) \in \{1, \dots, G\}$, where the score is the generated token.

\section{Proposed Approach: LLM-as-a-Verifier}
\subsection{Motivation}
\begin{wrapfigure}{r}{0.36\textwidth}
    \centering
    \vspace{-33pt}
    \includegraphics[width=\linewidth]{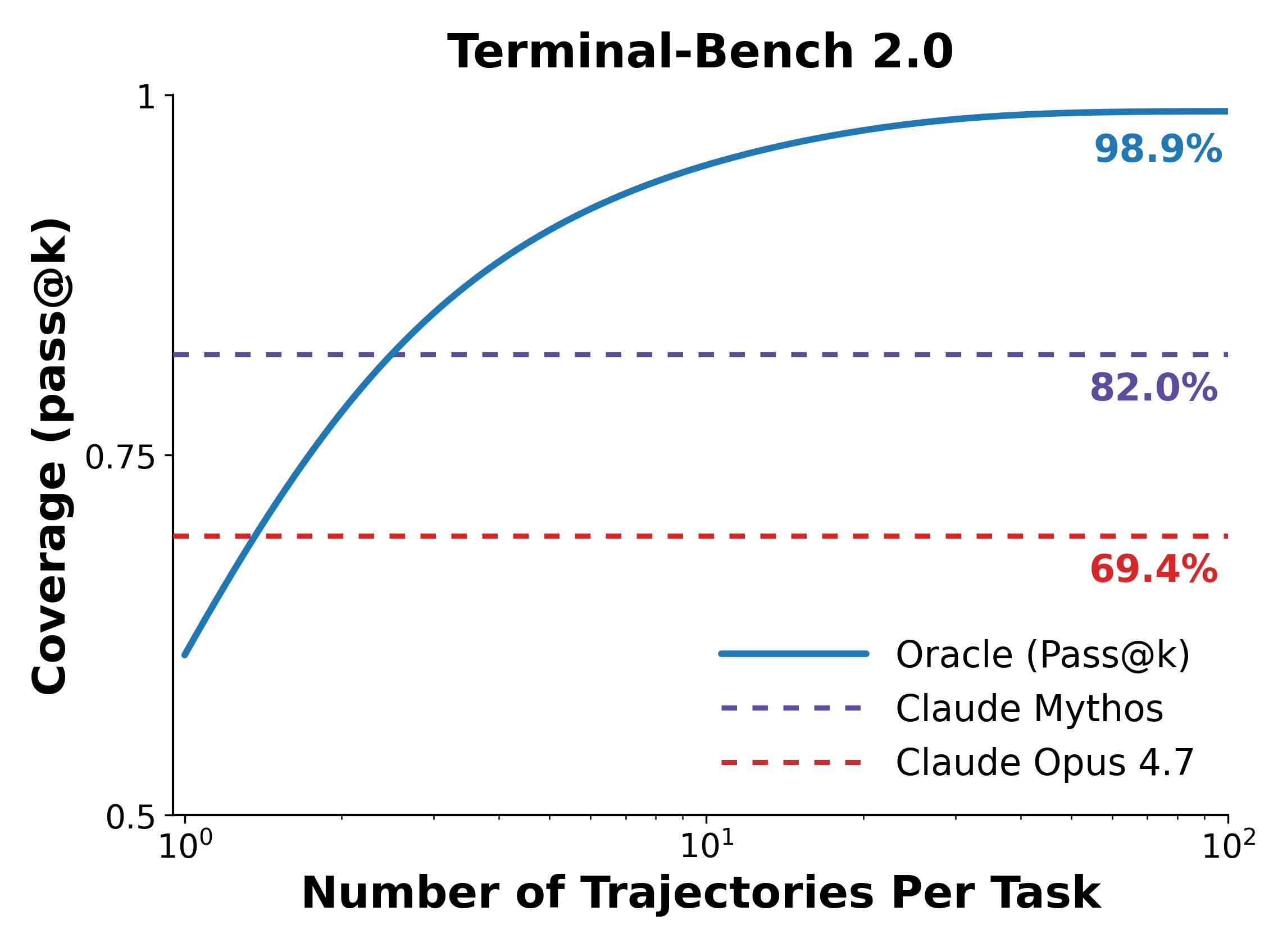}
    \vspace{-16pt}
    \caption{Oracle Pass@$K$ reaches 98.9\% on Terminal-Bench~V2.}
    \label{fig:motivation}
    \vspace{-8pt}
\end{wrapfigure}
\begin{figure}[t]
    \centering
\includegraphics[width=\linewidth]{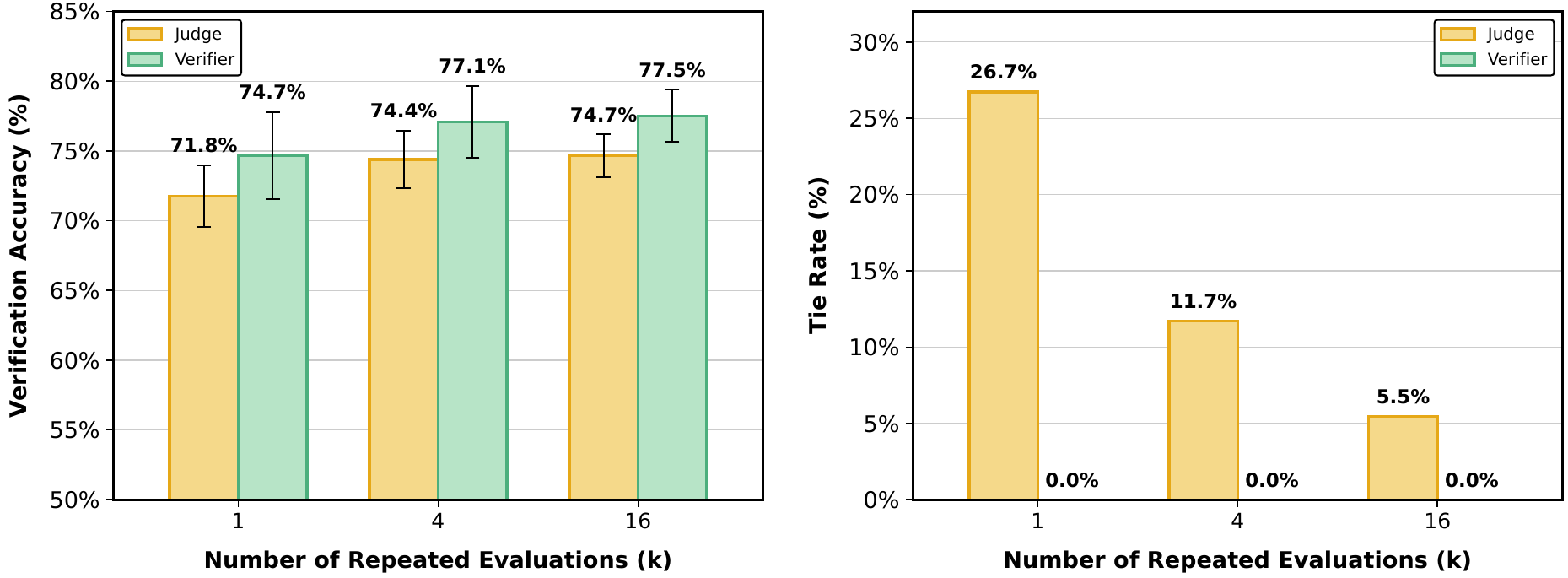}
    \caption{\textbf{Verifier (continuous) vs.\ Judge (discrete)} on Terminal-Bench~V2 across $k \in \{1, 4, 16\}$ repeated evaluations.
    \textbf{Left:} Pairwise verification accuracy. The verifier achieves $74.7\%$ at $k{=}1$ and improves to $77.5\%$ at $k{=}16$, consistently outperforming the judge across all evaluation budgets. 
    \textbf{Right:} Tie rate. The judge produces ties in $26.7\%$ of comparisons at $k{=}1$ due to coarse discrete scoring, decreasing to $5.5\%$ at $k{=}16$ as averaging breaks ties. In contrast, the verifier yields zero ties.}
    \label{fig:judge_verifier}
\end{figure}

Most models already possess the capability to solve many tasks: when executed repeatedly, they often produce a correct solution at least once. As shown in Fig.~\ref{fig:motivation} (left), the fraction of solved tasks increases consistently as we scale the number of sampled trajectories on Terminal-Bench, assuming access to an oracle verifier that always picks the optimal trajectory. Under this setting, the success rate reaches 98.9\% when pooling trajectories across the full Terminal-Bench~V2 leaderboard, effectively solving nearly the entire benchmark. However, capturing this headroom requires a verifier that can reliably distinguish correct trajectories from incorrect ones. While standard LM judges~\citep{zheng_judging_nodate} can be used as verifiers, they fail to provide sufficiently fine-grained feedback. Specifically, they prompt the model to output a discrete score token and select the highest-probability token as the final score, collapsing the full scoring distribution into a single value. This leads to inherently coarse evaluations. When comparing complex solutions, standard LM judges often assign the same score, resulting in ties and failing to discriminate between them. As a result, coarse scoring induces a high tie rate (27\%) on Terminal-Bench, with distinct trajectories often collapsing to the same score, as illustrated in Figure~\ref{fig:judge_verifier}. One could instead train a reward model~\citep{stiennon_learning_2022}, but such methods are constrained by their training data and often fail to generalize across domains. These limitations motivate the need for a generalizable framework that can provide fine-grained verification signals.

\subsection{Methodology}
\textbf{Fine-Grained Reward Estimation.} By definition, a judge is one who forms an overall opinion and assigns a decision, whereas a verifier is one who confirms the truth or correctness of something and requires more detailed evaluations. To this end, we introduce LLM-as-a-Verifier, a probabilistic verification framework that provides fine-grained feedback by scaling scoring granularity, repeated evaluation, and criteria decomposition. 

Let $V_{\text{score}} = \{v_1, \ldots, v_G\}$ denote an ordered set of tokens representing discrete score levels. Given a task prompt $x$, a language model $p_\theta$, a criterion $c$, and two candidate trajectories $\tau_i$ and $\tau_j$, we construct scoring prompts and obtain their conditional distributions $p_{\theta}(v \mid x,c,\tau_i)$ and $p_{\theta}(v \mid x, c, \tau_j)$ by extracting the logprobs from $<score_A>$ and $<score_B>$ tags using the following prompt:
\begin{quote}
\small
\ttfamily
You are an expert [domain] reviewer. You will see a task description and two trajectories.

Evaluation Criteria: [domain specific criteria]

Task: \{task prompt\} \\
Trajectory A: \{A\} Trajectory B: \{B\}

Carefully analyze each trajectory, then provide your final scores:
\begin{verbatim}
<score_A> INTEGER_1_TO_20 </score_A>
<score_B> INTEGER_1_TO_20 </score_B>
\end{verbatim}

Rating Rules: Rate correctness on a 1--20 scale based on evaluation criteria (1 = incorrect, 10 = borderline, 20 = correct)

Note: We use a letter-based scale instead of digits to enable logprob extraction for granularity scaling. 
\end{quote}

Rather than collapsing each distribution to a single discrete score, we approximate the reward of a trajectory as:
\begin{equation}
R(x, \tau)
= \frac{1}{CK} \sum_{c=1}^{C} \sum_{k=1}^{K}
\sum_{g=1}^{G} p_{\theta}(v_g \mid x, c, \tau)\,\phi(v_g)
\label{eq:main}
\end{equation}
where $C$ is the number of evaluation criteria, $K$ is the number of repeated verifications, $G$ is the number of score tokens (granularity level), $p_{\theta}(v_g \mid x, c, \tau)$ is the probability assigned by model $\theta$ to score token $v_g$, and $\phi(v_g)$ maps each score token to a scalar value.

We first normalize $R(x,\tau)\in[0,1]$ by the linear map $R\mapsto(R-\phi_{\min})/(\phi_{\max}-\phi_{\min})$. Then, we convert these continuous rewards into a pairwise preference using the Bradley--Terry model, treating $R(x,\tau)$ as the latent strength of trajectory $\tau$:
\begin{equation}
P(\tau_i \succ \tau_j \mid x)
\;=\; \frac{1}{1+\exp\!\big(-(R(x,\tau_i)-R(x,\tau_j))\big)},
\label{eq:pref}
\end{equation}

\begin{figure}[t]
    \centering
    \includegraphics[width=\textwidth]{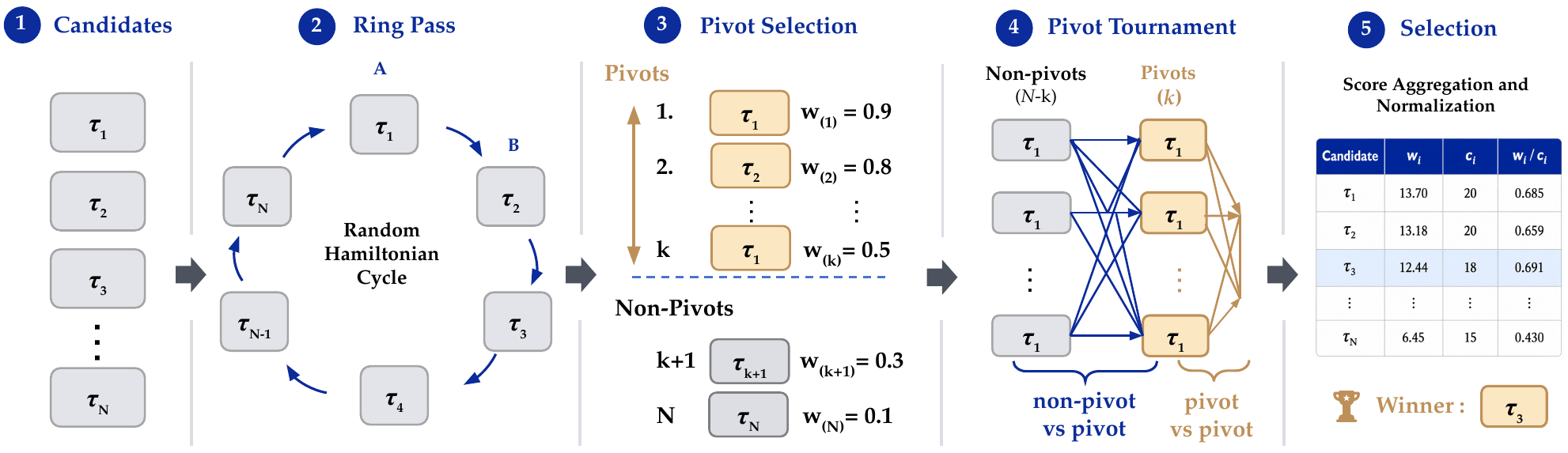}
    \caption{\textbf{Probabilistic Pivot Tournament.} A five-stage pipeline for selecting the best of $N$ candidates under a constrained verification budget.
    \textbf{(1)~Candidates:} the pool $\{\tau_1,\dots,\tau_N\}$ to be ranked.
    \textbf{(2)~Ring pass:} a random Hamiltonian cycle scores the $N$ adjacent pairs so every candidate appears once in the ``A'' slot and once in ``B'', canceling the model's positional bias.
    \textbf{(3)~Pivot selection:} candidates are ranked by their ring-pass scores $w_{(i)}$, and the top-$k$ candidates form the pivot set $\mathcal{P}$.
    \textbf{(4)~Pivot tournament:} every \emph{non-pivot--vs--pivot} and \emph{pivot--vs--pivot} pair is scored via Eq.~\ref{eq:pref}, concentrating the budget on uncertain top candidates and cutting cost from $\mathcal{O}(N^2)$ to $\mathcal{O}(Nk)$.
    \textbf{(5)~Selection:} comparisons are aggregated into win mass $w_i$ and count $c_i$, and the candidate with the highest normalized $w_i/c_i$ is returned.}
    \label{fig:pivot_tournament}
\end{figure}

\paragraph{Probabilistic Pivot Tournament.}
To pick the best trajectory among $N$ candidates, we can run a round-robin tournament that scores all $\binom{N}{2}$ pairs and accumulates wins
$$
w_i \;\mathrel{+}=\; P(\tau_i \succ \tau_j \mid x), \qquad
w_j \;\mathrel{+}=\; 1 - P(\tau_i \succ \tau_j \mid x),
\label{eq:soft-update}
$$
using the preference probability of Eq.~\ref{eq:pref}. However, such a schedule scales as $\mathcal{O}(N^2)$ pairwise verifications and quickly dominates verifier cost as $N$ grows. We propose a budget-efficient alternative, Probabilistic Pivot Tournament (PPT), illustrated in Fig.~\ref{fig:pivot_tournament}, in which every candidate is compared only against a small set of $k\!\ll\!N$ pivots, reducing the budget from $\mathcal{O}(N^2)$ to $\mathcal{O}(Nk)$. Critically, the choice of pivots determines whether the saved budget is well spent: arbitrary anchors waste verifications on candidates that are clearly weak. We therefore introduce a \emph{ring-based pivot selection} step that both removes the verifier's positional bias and concentrates the remaining budget on uncertain top candidates. PPT proceeds in three steps:

\textit{1) Ring pass.} We sample a uniformly random Hamiltonian cycle $\gamma$ over $\{1,\ldots,N\}$ and score the $N$ adjacent pairs
$\{(\gamma_t,\gamma_{t+1 \,\mathrm{mod}\, N})\}_{t=1}^{N}$. By the cyclic structure, every candidate appears \emph{exactly once in the ``A'' position and once in the ``B'' position} of the verifier prompt, so any systematic preference of the language models for one slot over the other cancels in expectation across the ring.

\textit{2) Pivot selection.} We rank candidates by their ring-pass mean preference $w_i/c_i$ and choose the top-$k$ as the pivot set $\mathcal{P}$. Selecting pivots from the empirical leaders allocates the remaining verification budget to the candidates most likely to be correct, so the subsequent pairwise comparisons distinguish among uncertain top candidates rather than spending queries on weak anchors.

\textit{3) Pivot rounds.} With the pivot set fixed, we score (i) every \emph{non-pivot vs.\ pivot} pair $(i,p)$ with $i\!\notin\!\mathcal{P},\,p\!\in\!\mathcal{P}$, and (ii) every \emph{pivot vs.\ pivot} pair within $\binom{\mathcal{P}}{2}$. All ring and pivot-round comparisons are aggregated into the same $w_i$, $c_i$, and we select $i^\star \in \arg\max_i\,w_i/c_i$. Normalizing by $c_i$ removes the bias that pivots participate in more comparisons than non-pivots. The total number of pairwise verifications is $N + k(N-k) + \binom{k}{2}$ which scales as $\mathcal{O}(Nk)$ where $k\!\ll\!N$. The full generation and verification pipeline is given in Algorithm~\ref{alg:testtime} (Appendix~\ref{app:ppt}).

To rigorously evaluate the ranking algorithms and assess performance on large candidate pools, we curate 20 trajectories per task using the Terminus-2 harness, and benchmark all methods in this setting. Table~\ref{tab:pivot-budget} characterizes the budget–accuracy trade-off of PPT, showing that our method outperforms prior approaches (e.g., V1~\cite{singh_v_1_2026}) while requiring fewer comparisons. Notably, performance improves consistently as the number of pivots increases. Further ablations in Appendix~\ref{app:ppt}.
\vspace{-1pt}
\section{Verification Scaling}
\label{sec:scaling}

Equation~\ref{eq:main} illustrates three independent axes along which verification can be scaled: the granularity of score tokens $G$, the number of repeated evaluations $K$, and the number of evaluation criteria $C$. Each axis targets a different source of error in the reward estimate, and we find that the three act as complementary levers: increasing granularity improves score separation between candidate solutions, repeated evaluation averages out biases from individual verification passes, and criteria decomposition captures complementary aspects of trajectory quality. For all scaling experiments, we use Gemini~2.5~Flash~\citep{gemini25flash} as the verifier, which allows us to extract up to 20 top logprobs per scoring token. In Fig.~\ref{fig:scaling_verifier}, we show that verification accuracy on Terminal-Bench 2.0 improves along all three dimensions, rising from $73.1\%$ at $G{=}1$ to $77.5\%$ at $G{=}20$, from $74.7\%$ at $K{=}1$ to $77.4\%$ at $K{=}16$, and from $75.2\%$--$76.4\%$ for any single criterion to $78.3\%$ when the three criteria are ensembled. We measure the pairwise verification accuracy over 200 randomly sampled trajectories from Terminal-Bench, spanning multiple agent harnesses. Each axis is a knob that the practitioner can tune depending on the latency budget of the downstream application. While our primary experiments use a logprob-accessible model, Appendix~\ref{app:closed-api} demonstrates that our framework is also compatible with frontier models that do not expose token-level log-probabilities via a simple two-stage workaround.

\subsection{Scoring Token Granularity}
\label{sec:granularity}

Standard LM judges collapse the scoring distribution to the single highest-probability token, yielding a discrete reward $R_{\text{LM}}(t,\tau)\in\{1,\ldots,G\}$ with resolution $1/G$. Intuitively, enlarging the ordered token set $V_{\text{score}}$ does not grant the verifier any new information about the trajectory. Yet, it grants the decoder a finer space in which to project the model's internal belief, so that nearby beliefs that would have been rounded to the same integer are now mapped to continuous rewards.

\begin{table}[h!]
    \centering
    \begin{minipage}{0.40\linewidth}
        \centering
        \begin{equation}
        \mathrm{SNR}(G) = \frac{\mathbb{E}[s_c - s_i]}{\sqrt{\mathrm{Var}(s_c - s_i)}}
        \label{eq:snr}
        \end{equation}
    \end{minipage}\hfill
    \begin{minipage}{0.56\linewidth}
        \centering
        \small
        \begin{tabular}{ccccc}
        \toprule
        Granularity $G$ & 1 & 4 & 16 & 20 \\
        \midrule
        SNR ($k{=}16$) & 0.775 & 0.786 & 0.797 & \textbf{0.799} \\
        \bottomrule
        \end{tabular}
    \end{minipage}
    \caption{\textbf{Signal-to-noise ratio (SNR).} \emph{(Left)} The SNR measures how reliably the verifier separates correct ($s_c$) from incorrect ($s_i$) trajectories (Eq.~\ref{eq:snr}). \emph{(Right)} As the number of scoring tokens $G$ increases, the SNR grows, indicating better-calibrated score separation.}
    \label{tab:snr}
\end{table}

\paragraph{Signal-to-Noise Ratio.} To isolate why finer granularity improves verification, we decompose the pairwise score gap $\Delta = s_c - s_i$ between correct ($s_c$) and incorrect ($s_i$) trajectories into a signal and a noise component. We define the signal-to-noise ratio as in Eq.~\ref{eq:snr} (Table~\ref{tab:snr}, left), where $\mathbb{E}(s_c-s_i)$ captures how strongly the verifier prefers the correct trajectory over the incorrect one (\emph{signal strength}), and the denominator, $\mathrm{Var}(s_c - s_i)$, captures how inconsistent that preference is across pairs (\emph{noise}). Pairwise verification accuracy is a monotonic function of $\mathrm{SNR}(G)$: holding sample size fixed, a larger standardized gap implies a higher probability that $s_c>s_i$. Empirically, we find that $\mathrm{SNR}(G)$ increases from $0.775$ at $G{=}1$ to $0.799$ at $G{=}20$ on Terminal-Bench (Table~\ref{tab:snr}). Finer-grained tokens therefore produce better-calibrated scores that more reliably separate correct from incorrect trajectories, which in turn improves the pairwise accuracy from $73.1\%$ to $77.5\%$.

\vspace{-0.15in}
\paragraph{Case Study: \texttt{query-optimize}.}
To concretely illustrate how scaling granularity to $G{=}20$ and our probabilistic formulation sharpen the verifier's signal, we analyze a representative trajectory pair from the \texttt{query-optimize} task on Terminal-Bench~V2, generated by Claude~Opus~4.5 under the OpenHands harness and scored by Gemini~2.5~Flash. Here the agent is given a slow SQL query over a database and asked to produce an equivalent optimized version. Both candidate trajectories generate queries that execute faster, but they differ critically in their verification procedures. The correct trajectory waits the full $5$ minutes for the original query to complete on the canonical database and performs a direct \texttt{diff} against the optimized output. In contrast, the failing trajectory never validates equivalence on the database and instead creates a new database. As shown in the reasoning traces in Appendix~\ref{app:query-optimize-example}, Gemini~2.5~Flash reliably identifies this failure mode, but expresses it in graded, hedged language (e.g., \textit{``slightly cleaner,''} \textit{``marginally more direct''}), as if the discrepancy were minor. When evaluated over 100 repetitions, a standard LM judge on a $1$--$5$ scale collapses these nuanced assessments into discrete scores (Table~\ref{tab:query-optimize-main}), producing ties (e.g., $5$ vs.\ $5$) in $88$ out of $100$ runs, thus failing to meaningfully discriminate between the candidates. Taking the expectation over the \emph{same} $5$-point distribution eliminates ties entirely---ranking the correct trajectory higher in $69$ runs---and scaling the granularity to $G{=}20$ sharpens the signal further, letting LLM-as-a-Verifier rank the correct trajectory strictly higher in $77$ out of $100$ runs.

\begin{table}[h!]
    \centering
\caption{\textbf{Judges vs.\ Verifiers on \texttt{query-optimize}.} Over $100$ repeated evaluations, we count how often the correct trajectory is scored higher than ($s_c{>}s_i$), tied with ($s_c{=}s_i$), or lower than ($s_c{<}s_i$) the incorrect one. The discrete $1$--$5$ judge produces ties in $88/100$ evaluations. Taking the expectation over the same $1$--$5$ scale eliminates ties and correctly ranks the trajectory in $69/100$ evaluations. Increasing the score granularity to $G{=}20$ further improves discrimination, correctly ranking the trajectory in $77/100$ evaluations.}
    \label{tab:query-optimize-main}
    \small
    \renewcommand{\arraystretch}{1.3}
    \begin{tabular}{llccc}
    \toprule
    Method & Reward $R(x,\tau)$ & $\#(s_c{>}s_i)$\,{\color{deepgreen}\cmark} & $\#(s_c{=}s_i)$~\textbf{\textit{(tie)}} & $\#(s_c{<}s_i)$\,{\color{deepred}\xmark} \\
    \midrule
    Judge (discrete, $G{=}5$)        & $\phi\!\left(\arg\max_g p_{\theta}(v_g)\right)$       & $12/100$ & $88/100$ & $0/100$ \\
Verifier (continuous, $G{=}5$)
& $\sum_{g=1}^{5} p_{\theta}(v_g)\,\phi(v_g)$
& $69/100$ & $0/100$ & $31/100$ \\

\textbf{Verifier (continuous, $G{=}20$)}
& $\sum_{g=1}^{20} p_{\theta}(v_g)\,\phi(v_g)$
& $\mathbf{77/100}$ & $\mathbf{0/100}$ & $23/100$ \\
    \bottomrule
    \end{tabular}
\end{table}

\subsection{Repeated Evaluation}
\label{sec:repeated}

While granularity improves score calibration within a single forward pass, it does not address a second source of error: the verifier's variance on one evaluation. Even at high $G$, a single evaluation $R^{(k)}(x,\tau)$ can be skewed by spurious features of the prompt or failure modes of the verifier on a particular trajectory. Averaging $K$ independent evaluations $\frac{1}{K}\sum_{k=1}^{K} R^{(k)}(x,\tau)$ is a Monte Carlo estimator of the underlying expected reward; its variance shrinks as $\mathcal{O}(1/K)$ while its bias is unchanged. This complements granularity rather than duplicating it: granularity sharpens each individual estimator, while repeated evaluation averages out the noise that granularity cannot remove. Fig.~\ref{fig:scaling_verifier} (middle) shows that the accuracy increases from $74.7\%$ at $K{=}1$ to $77.5\%$ at $K{=}16$. However, gains diminish with larger $K$: early improvements arise from variance reduction, while additional evaluations contribute diminishing returns due to correlated biases on harder examples. Importantly, repeated evaluation benefits discrete judges, whose coarse scoring induces high tie rates at low $K$. While increasing $K$ helps break these ties through averaging, this mechanism is fundamentally limited for discrete judges. We show that a single-pass verifier ($K{=}1$) already matches a heavily ensembled judge ($K{=}16$), highlighting that fine-grained probabilistic scoring provides a stronger signal.

\subsection{Criteria Decomposition}
\label{sec:criteria}

Granularity and repeated evaluation both assume that the rubric itself is adequate; neither helps if a single monolithic criterion is a poor proxy for trajectory quality. In long-horizon agentic tasks, judgments like ``is this trajectory correct?'' conflate several logically distinct factors, and verifiers asked a compound question often latch onto whichever factor is most salient in the prompt. Therefore, we replace a single monolithic rubric with an ensemble over $C$ simpler sub-criteria. Concretely, for code-agent trajectories we decompose correctness into three factors that are individually easier to verify: Specification (whether the trajectory satisfies all task requirements), Output (whether the final output format matches the expected result), and Errors (whether the trajectory is free of failure signals in logs and tool outputs). The final reward averages the expected scores across criteria, as in the outer sum of Eq.~1. In Fig.~\ref{fig:scaling_verifier} (right), any one criterion alone achieves $75.2\%$--$76.4\%$ accuracy, and their ensemble reaches $78.3\%$.
\section{Experiments}
\label{sec:experiments}

We evaluate LLM-as-a-Verifier as a trajectory reward model (TRM) for test-time scaling across four benchmarks that span three domains: coding (Terminal-Bench V2~\citep{merrill_terminal-bench_2026}, SWE-Bench Verified~\citep{jimenez_swe-bench_2024}), robotics (RoboRewardBench~\citep{lee2026roboreward}), and medical (MedAgentBench~\citep{jiang2025virtual}). Across all four, we use the same protocol: a generation policy $\pi_\theta$ produces $N$ candidate trajectories per task, the verifier scores every pair using the probabilistic pivot tournament as described in Algorithm~\ref{alg:testtime}, and the trajectory with the highest normalized score is submitted. Unless otherwise noted, the verifier is run with granularity $G{=}20$, repeated evaluations $K{=}8$, and the three-criterion decomposition described in Section~\ref{sec:criteria}. Our method is training-free and plug-and-play: the same verification framework is applied across all four benchmarks without any per-domain fine-tuning. Overall results are summarized in Fig.~\ref{fig:teaser}, and per-benchmark headline numbers, including baseline accuracies, Pass@1, and oracle Pass@$N$, are reported in Table~\ref{tab:per-bench-headline}.

\begin{table}[h]
\centering
\caption{\textbf{Per-benchmark performance and gains from verification.} Baseline accuracies (left) are obtained under a fixed agent harness. On the same candidate pools, we report Pass@1, the oracle Pass@$N$ upper bound, and the accuracy achieved by LLM-as-a-Verifier (right). Our method consistently improves over Pass@1 and recovers a large portion of the oracle headroom, achieving state-of-the-art performance on each.}
\label{tab:per-bench-headline}
\small
\setlength{\tabcolsep}{4pt}
\begin{tabular}{l ccc ccc}
\toprule
 & \multicolumn{3}{c}{\textbf{Baseline models (accuracy)}} & \multicolumn{3}{c}{\textbf{LLM-as-a-Verifier}} \\
\cmidrule(lr){2-4}\cmidrule(lr){5-7}
Benchmark & \#1 & \#2 & \#3 & Pass@1 & Oracle & \textbf{Ours} \\
\midrule
Terminal-Bench V2  & GPT-5.5 (84.7\%)        & Opus 4.7 (80.2\%) & G3.1 Pro (80.2\%) & 83.1\% & 92.1\% & \textbf{86.5\%} \\
SWE-Bench Verified & Opus 4.5 (76.8\%)       & G3 Flash (75.8\%)        & M2.5 (75.8\%)           & 76.1\% & 84.4\% & \textbf{78.2\%} \\
MedAgentBench      & Opus 4.8 (70.2\%)       & G3.5 Flash (66.3\%)        & GPT-5.5 (65.1\%)       & 70.2\% & 75.0\% & \textbf{73.3\%} \\
\bottomrule
\end{tabular}
\end{table}

\subsection{Terminal-Bench V2}
\label{sec:terminal}

Terminal-Bench V2~\citep{merrill_terminal-bench_2026} measures an agent's proficiency in shell-based environments across long-horizon tasks that require multi-step reasoning, file manipulation, and recovery from failed tool calls. The benchmark is particularly difficult for verifiers because many trajectories produce syntactically plausible but incorrect terminal states. We use Capy~\citep{capy} as the scaffold and sample $N{=}5$ trajectories per task from GPT-5.5; Gemini 2.5 Flash serves as the verifier. The Pass@1 of GPT-5.5 under Capy is $83.1\%$, and the oracle Pass@5 upper bound on this candidate pool is $92.1\%$. LLM-as-a-Verifier improves the accuracy from $83.1\%$ to $\mathbf{86.5\%}$, surpassing Claude Mythos + Terminus-2~\citep{noauthor_terminal-benchterminal_benchagentsterminus_2_nodate} ($82.0\%$), GPT-5.5 + NexAU-AHE ($84.7\%$), Claude Opus 4.7 + WOZCODE ($80.2\%$), and Gemini 3.1 Pro + TongAgents ($80.2\%$) and setting a new state of the art on Terminal-Bench V2.\footnote{Baseline accuracies are extracted from the official Terminal-Bench~V2 and SWE-Bench Verified leaderboard.} We further show that these gains are not tied to a specific harness. For additional generalization results on Terminus-2 and Terminus-Kira, refer to Appendix~\ref{app:harness}.

\subsection{SWE-Bench Verified}
\label{sec:swebench}

SWE-Bench Verified~\citep{jimenez_swe-bench_2024} is a human-curated subset of 500 real-world GitHub issues where each task requires an agent to produce a patch that resolves the issue and passes the maintainer's hidden test suite. It stresses long-context reasoning, cross-file edits, and compliance with an existing codebase. We use mini-swe-agent as the scaffold and, in contrast to the homogeneous proposal pool used on Terminal-Bench, draw a heterogeneous pool of $N{=}3$ candidates per task by sampling one trajectory each from Claude Opus 4.5, Gemini 3 Flash, and MiniMax M2.5. Gemini 2.5 Flash again serves as the verifier. The mean Pass@1 across this candidate pool is $76.1\%$ and the oracle Pass@3 upper bound is $84.4\%$. LLM-as-a-Verifier achieves $\mathbf{78.2\%}$ on SWE-Bench Verified, outperforming Claude Opus 4.5 ($76.8\%$), Gemini 3 Flash ($75.8\%$), and MiniMax M2.5 ($75.8\%$). These results highlight the verifier's ability to select the strongest trajectory from a diverse set of candidates produced by different model families.

\subsection{RoboRewardBench}
\label{sec:roboreward}

\begin{wraptable}{r}{0.42\textwidth}
\vspace{-\baselineskip}
\centering
\caption{\textbf{Preference accuracy on RoboRewardBench.} LLM-as-a-Verifier outperforms trained robotics reward models.}
\label{tab:roboreward}
\small
\setlength{\tabcolsep}{6pt}
\begin{tabular}{lc}
\toprule
Method & Accuracy (\%) \\
\midrule
TOPReward                  & 74.7 \\
Robometer-4B               & 78.8 \\
RoboReward-8B              & 81.4 \\
LLM-as-a-Judge (Discrete)  & 70.8 \\
\textbf{LLM-as-a-Verifier (Ours)} & \textbf{87.4} \\
\bottomrule
\end{tabular}
\vspace{-\baselineskip}
\end{wraptable}

RoboRewardBench~\citep{lee2026roboreward} evaluates reward models on robotic manipulation trajectories. Following~\citet{liang2026robometer}, we curate pairs of rollout videos that follow the same natural-language instruction but make different amounts of progress; the reward model must output a preference indicating which rollout makes more progress. Unlike the coding and clinical benchmarks, inputs here are multi-frame videos, so the verifier must integrate visual context across frames to reason about physical progress toward the goal. We use Qwen 3.6 35B as the base VLM verifier and apply the same probabilistic formulation (Eq.~1) over scoring tokens extracted from the VLM's logits, with granularity $G{=}20$ and $K{=}8$ repeated verifications. We evaluate on 500 randomly sampled trajectory pairs from the RoboRewardBench and compare against (i)~a discrete LLM-as-a-Judge baseline using the same VLM, (ii)~reward models specifically trained on robotics data---RoboReward-8B (trained on $\sim$45k episodes) and Robometer-4B (trained on $\sim$1M comparisons), and (iii)~TOPReward~\cite{chen2026topreward} (Qwen 3.6). As shown in Table~\ref{tab:roboreward} and Appendix~\ref{app:roboreward}, LLM-as-a-Verifier achieves $\mathbf{87.4\%}$ preference accuracy, outperforming the discrete LLM-as-a-Judge baseline ($70.8\%$), RoboReward-8B ($81.4\%$), Robometer-4B \cite{liang2026robometer} ($78.8\%$), and TOPReward (74.7\%), despite being applied zero-shot and without any fine-tuning. We also evaluate on RoboRewardBench by measuring the Mean Absolute Error (MAE) between predicted rewards and human annotations. As shown in Table~\ref{tab:roboreward-mae}, using the continuous reward formulation in Eq.~\ref{eq:main} together with $K{=}8$ repeated evaluations substantially improves alignment with human judgments, reducing the MAE from $1.11$ to $\mathbf{0.72}$.

\begin{table}[h]
\centering
\caption{\textbf{Evaluation on RoboRewardBench against human annotations.} We report Mean Absolute Error (MAE; lower is better) between human labels and predicted rewards. LLM-as-a-Verifier uses continuous rewards ($K{=}8$), whereas the baseline extracts discrete scores from the model.}
\label{tab:roboreward-mae}
\small
\setlength{\tabcolsep}{8pt}
\begin{tabular}{lc}
\toprule
Models & RoboRewardBench MAE ($\downarrow$) \\
\midrule
RoboReward 8B                          & 1.11 \\
\textbf{RoboReward 8B + LLM-as-a-Verifier} & \textbf{0.72} \\
\bottomrule
\end{tabular}
\end{table}

\subsection{MedAgentBench}
MedAgentBench~\citep{jiang2025virtual} evaluates LLM agents on medical tasks that involve patient information retrieval, guideline lookup, and multi-step tool use in a simulated electronic health records (EHR) environment. It covers a regime where ground-truth trajectory checkers are expensive to construct and where verification errors carry real safety consequences, making it a natural stress test for general-purpose verifiers. We use the AgentBench harness and sample $N{=}5$ trajectories per task from Claude Opus 4.8, then apply the same verification procedure. The Pass@1 of Claude Opus 4.8 on this pool is $70.2\%$ and LLM-as-a-Verifier achieves $\mathbf{73.3\%}$, outperforming Opus 4.8 ($70.2\%$), Gemini 3.5 Flash ($66.3\%$), and GPT-5.5 ($65.1\%$).

\begin{figure}[t!]
    \centering
    \includegraphics[width=0.98\linewidth]{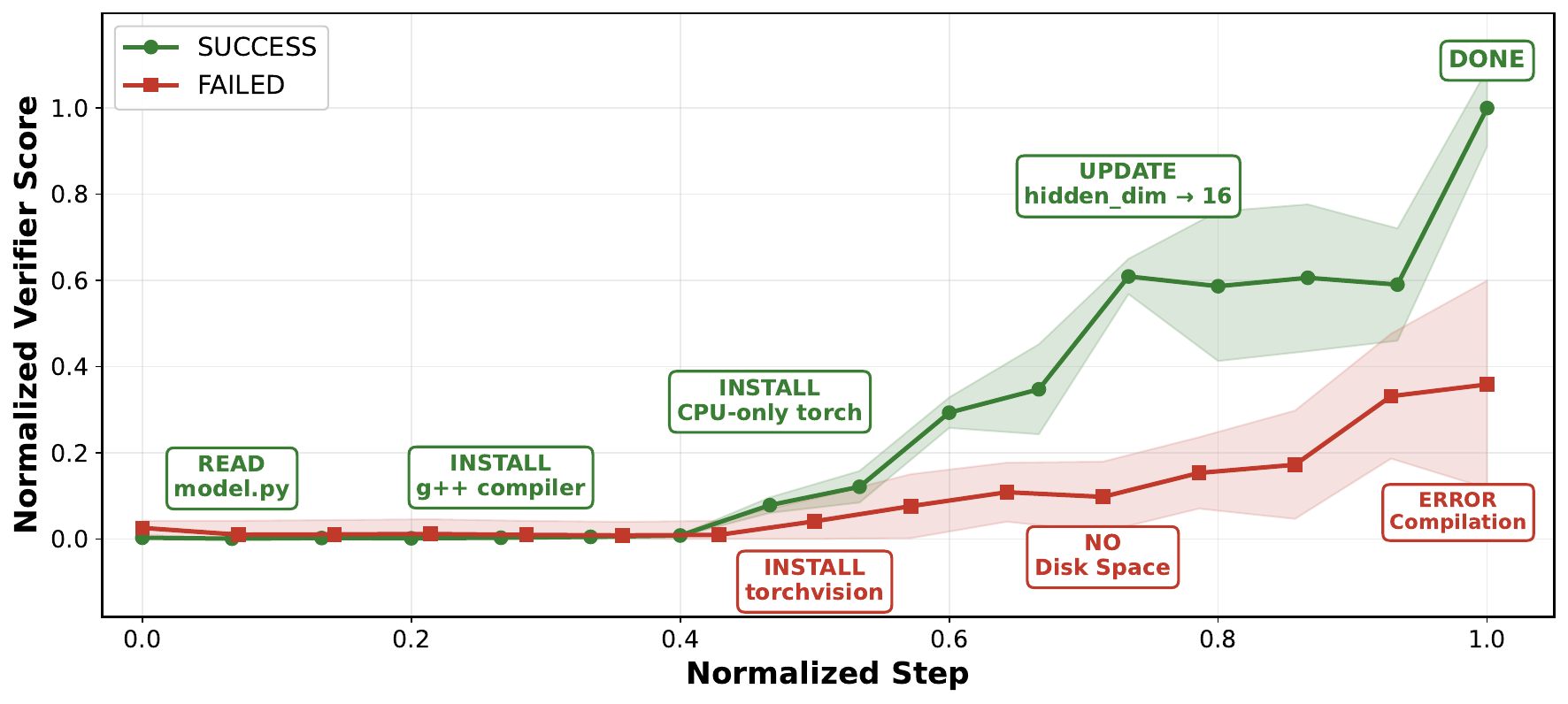}
    \caption{We observe a strong correlation between the chronological progression of code generation steps and the scores from LLM-as-a-Verifier. The example task above requires the agent to run MNIST inference. The successful trajectory follows a coherent sequence of events---\emph{Read \texttt{model.py} $\rightarrow$ Install g++ compiler $\rightarrow$ Install CPU-only torch $\rightarrow$ Update \texttt{hidden\_dim}} $\rightarrow$ DONE and exhibits consistently increasing verifier scores. In contrast, the failed trajectory is characterized by erroneous behaviors---it unnecessarily installs the large \texttt{torchvision} package, which exhausts the available disk space and hits a compilation error---resulting in significantly lower scores. Results are shown for the \texttt{pytorch-model-cli} task from Terminal-Bench V2, using Gemini 2.5 Pro with Terminus 2 and Gemini 2.5 Flash as the verifier.}
    \label{fig:voc-counterfactual-code}
\end{figure}

\section{Fine-grained Verifier Signals as a Proxy for Task Progress}
\label{sec:progress}

Beyond selecting the best trajectory, the fine-grained signal produced by LLM-as-a-Verifier can serve as a scalar proxy for how far an agent has progressed through a task. We quantify this using the \emph{Value-Order Correlation} (VOC), the Spearman rank correlation between the chronological index of a step and the verifier's predicted value for the prefix ending at that step, following \citet{ICLR2025_54854cf1}. Intuitively, a verifier that tracks task progress should assign monotonically higher scores to later prefixes of a successful rollout, yielding $\mathrm{VOC}\to 1$, and should remain robust to failure modes such as getting stuck or regressing.
\begin{equation}
\mathrm{VOC} = \mathrm{rank\text{-}correlation}\!\left(
\mathrm{argsort}(s_{t_1}, s_{t_2}, \cdots, s_{t_K}),
(t_1, t_2, \cdots, t_K)
\right).
\end{equation}

\paragraph{VOC on code generation.} On Terminal-Bench V2, we measure VOC between the chronological step of each agent action and the verifier's score on the corresponding trajectory prefix. LLM-as-a-Verifier produces consistently increasing scores on successful rollouts while remaining largely flat on trajectories that stall or drift toward failure, allowing the same scalar to serve as both a progress measure and an early-warning signal. Figure~\ref{fig:voc-counterfactual-code} illustrates this on the \texttt{pytorch-model-cli} task, where the successful run's score rises monotonically while the failed run's stays low. This dual use motivates our Claude Code and Codex extensions, which surface the live verifier score to the user so that long-running agentic jobs can be monitored, paused, or rolled back before they commit broken state to disk. Across 500 (success, failure) pairs drawn from Terminal-Bench V2 runs, the verifier (Gemini 2.5 Flash, $G{=}20$) attains Spearman VOC $0.848$ on successful trajectories and $0.769$ on failed ones (full numbers reported in Table~\ref{tab:voc-terminal-by-outcome}). The code-generation VOC numbers indicate that the fine-grained signal produced by LLM-as-a-Verifier is not merely a better ranker but a calibrated estimator of task progress, opening a path toward safer real-world deployment of autonomous agents.

\begin{table}[h]
    \centering
    \begin{tabular}{lc}
        \toprule
        \textbf{Trajectory outcome} & \textbf{Spearman VOC (rank correlation)} \\
        \midrule
        Successful                                & $0.848 \pm 0.012$ \\
        Failed                                    & $0.769 \pm 0.016$ \\
        \midrule
        Success $-$ Failed (gap)                  & $+0.079$          \\
        \bottomrule
    \end{tabular}
\caption{\textbf{Value-Order Correlation by trajectory outcome on Terminal-Bench V2.}
Mean Spearman rank correlation between step index and verifier progress score, computed over 500 randomly sampled trajectories from Terminal-Bench V2. The verifier (Gemini 2.5 Flash, $G{=}20$) exhibits near-monotonic progress for successful trajectories, while failed rollouts show weaker correlation, indicating limited or inconsistent progress. We observe a $0.08$ Spearman gap between successful and failed trajectories generated by the same agent backbone on the same task.}
    \label{tab:voc-terminal-by-outcome}
\end{table}

\paragraph{VOC on robotics.} We compute VOC over 500 trajectories from the held-out RoboReward dataset. As shown in Table~\ref{tab:voc-roboreward}, LLM-as-a-Verifier (Qwen 3.6, $K{=}5$, $G{=}20$) attains $\mathbf{0.966}$, substantially exceeding RoboReward-8B ($0.877$), Robometer-4B ($0.780$), and TOPReward ($0.565$). Qualitatively, TOPReward tends to saturate at $P(\texttt{True}){=}1.0$ almost immediately and therefore loses the ability to discriminate mid-trajectory progress when a rollout eventually fails, whereas our expectation over the full scoring distribution preserves a smooth, chronologically-aligned signal throughout the episode.

\begin{table}[h]
    \centering
    \begin{tabular}{lc}
        \toprule
        \textbf{Method} & \textbf{Spearman VOC (rank correlation)} \\
        \midrule
        LLM-as-a-Verifier (Qwen 3.6 35B, 5 reps, 20 granularity) & \textbf{0.966} \\
        RoboReward-8B & 0.877 \\
        Robometer-4B & 0.780 \\
        TOPReward (Qwen 3.6, $P(\texttt{true})$) & 0.565 \\
        \bottomrule
    \end{tabular}
    \caption{\textbf{Value-Order Correlation on 500 trajectories from RoboRewardBench.} LLM-as-a-Verifier with $K{=}5$ repeated evaluations and $G{=}20$ scoring granularity attains the highest rank-correlation between the chronological step index and the verifier's predicted progress score.}
    \label{tab:voc-roboreward}
\end{table}

\paragraph{Coding Agent Extension} To demonstrate the applicability of LLM-as-a-Verifier to real-world coding agents, we develop \textbf{\textit{TurboAgent}}, a drop-in extension for Claude Code and other OpenAI-API compatible clients. TurboAgent operates as an inference-time proxy that transparently sits between the client and the LLM provider, requiring no modifications to either the underlying agent harness or the backend model. The proxy design also allows TurboAgent to be plugged transparently into existing benchmarks such as Terminal-Bench~\citep{merrill_terminal-bench_2026}. For each request, it dispatches $N$ candidate trajectories to the backend model in parallel and selects the best response using the proposed \emph{Probabilistic Pivot Tournament} (PPT). Beyond verification, TurboAgent also provides a web-based interface for visualizing verifier outputs and monitoring agent progress in real time.

\section{Dense Reward for Reinforcement Learning}
\label{sec:rl}

\begin{figure}[h!]
    \centering
    \includegraphics[width=\linewidth]{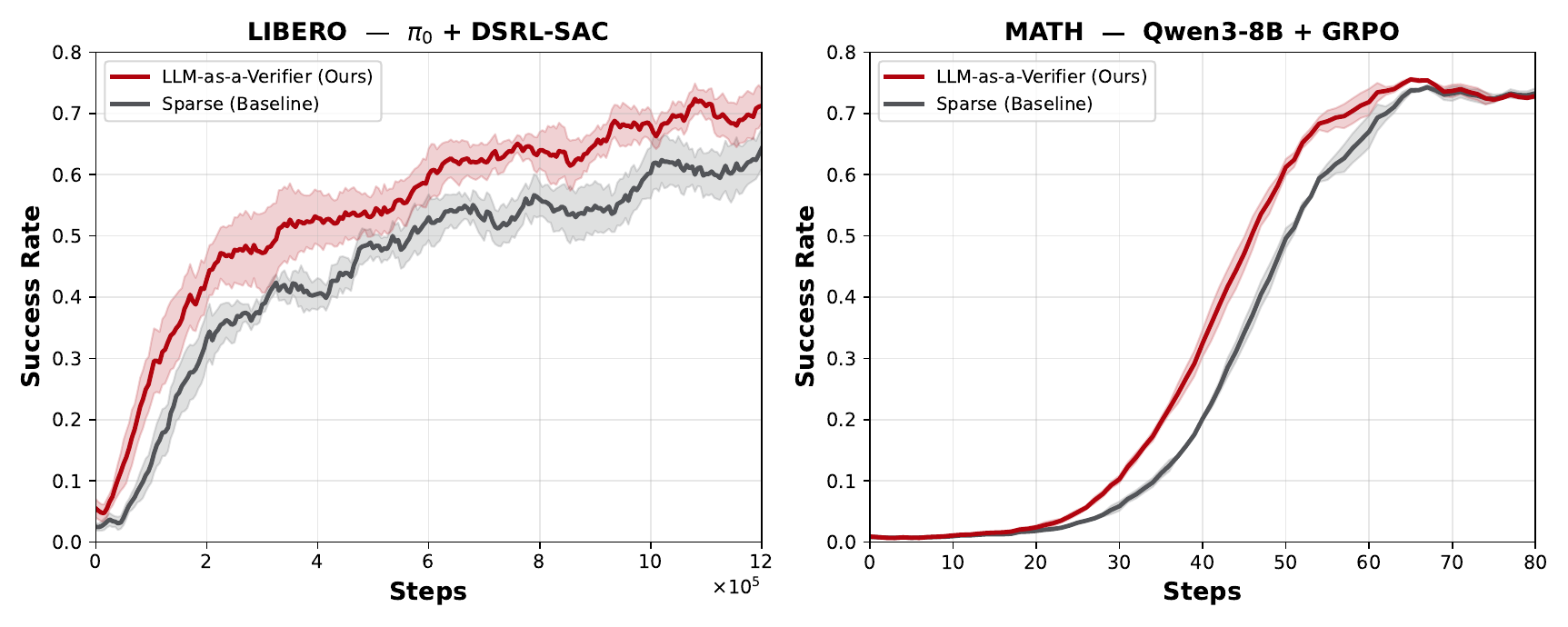}
    \caption{\textbf{LLM-as-a-Verifier improves RL sample efficiency.} Success rate versus training steps for off-policy (left) and on-policy (right) reinforcement learning, comparing sparse-reward baselines with dense rewards from LLM-as-a-Verifier. \textbf{Left:} A $\pi_0$ policy fine-tuned on the LIBERO \texttt{ketchup} task with DSRL-SAC. The verifier progress reward (Eq.~\ref{eq:rl-sac}) achieves the same success rate using $\approx1.8\times$ fewer environment steps and reaches a higher final success rate ($0.76$ vs.\ $0.69$). \textbf{Right:} Qwen3-8B fine-tuned on MATH with GRPO. The verifier reasoning reward (Eq.~\ref{eq:rl-grpo}) improves sample efficiency by $\approx1.1\times$. Results are averaged over multiple seeds (LIBERO $n{=}5$, MATH $n{=}3$).} 
    \label{fig:rl-sample-efficiency}
\end{figure}

The progress signal from the previous section also helps remediate a long-standing difficulty in reinforcement learning (RL): the \emph{credit assignment problem}. We show that the fine-grained score of LLM-as-a-Verifier (Eq.~\ref{eq:main}) is a drop-in dense reward for both off-policy and on-policy RL, improving sample efficiency without any reward-model training or environment-specific shaping.

\paragraph{Off-policy RL: dense progress rewards for DSRL-SAC.} We fine-tune the $\pi_0$~\citep{black2026pi0visionlanguageactionflowmodel} vision–language–action model on LIBERO with DSRL using Soft Actor–Critic (SAC). At the end of each rollout we query the VLM verifier with the task instruction $x$ and a uniformly sub-sampled sequence of rendered frames, obtaining a per-step progress curve $\rho_t = R\!\left(x, \tau_{1:t}\right)\in[0,1]$. We then relabel the rollout with the shaped reward:
\begin{equation}
r_t \;=\; r^{\text{env}}_t \;+\; \lambda\,\rho_t,
\label{eq:rl-sac}
\end{equation}
store the relabeled transitions $(s_t, a_t, r_t, s_{t+1})$ in the replay buffer $\mathcal{D}$, and train the SAC critic on the relabeled returns sampled from $\mathcal{D}$. The coefficient $\lambda$ trades off environment and verifier rewards. Because shaping is applied offline to stored trajectories and leaves the SAC objective untouched, it adds dense intermediate signals at no additional algorithmic cost.

\paragraph{On-policy RL: dense reasoning rewards for GRPO.} We fine-tune Qwen3-8B~\citep{yang2025qwen3} on MATH using Group Relative Policy Optimization (GRPO)~\citep{deepseek-ai_deepseek-r1_2025}, which samples a group of $G$ responses $\{y_i\}_{i=1}^{G}$ for each prompt $x$ and estimates each response's advantage relative to the group. During the early stages of training, it is common for all sampled responses to produce incorrect final answers, causing the group-relative advantage to collapse to zero and yielding no gradient. LLM-as-a-Verifier mitigates this issue by evaluating the reasoning trace of each completion using the probabilistic pivot tournament (Eq.~\ref{eq:pref}), assigning each response a normalized preference score $\bar{R}_i \in [0,1]$ that captures fine-grained differences in reasoning quality even when the final answers are identical. We incorporate this verifier-derived score into the standard correctness and format reward with weight $\beta$:
\begin{equation}
r_i
=
r_{\mathrm{correct},i}
+
r_{\mathrm{format},i}
+
\beta\, r_{\mathrm{reasoning},i}.
\label{eq:rl-grpo}
\end{equation}

\paragraph{Empirical findings.} Across both regimes, dense verifier rewards improve sample efficiency over the sparse baselines
(Fig.~\ref{fig:rl-sample-efficiency}). We quantify sample efficiency as the ratio of training steps the sparse baseline requires to reach a target success rate to the steps our dense reward requires. On LIBERO, shaping a $\pi_0$ policy trained with DSRL-SAC reaches a matched success rate in substantially fewer environment steps--- $\mathbf{1.8\times}$ higher sample efficiency across success-rate targets from $0.2$ to $0.6$---while also attaining a higher final success rate ($0.76$ vs.\ $0.69$). On MATH, augmenting GRPO with the reasoning reward gives a smaller but consistent gain of $\approx\mathbf{1.1\times}$ (a $\sim\!10\%$ reduction in the optimizer steps needed to reach a matched accuracy). We report reward-shaping hyperparameters and additional ablations in Appendix~\ref{app:rl}.
\section{Discussion}
\label{sec:conclusion}

In this work, we argue that verification constitutes an underexplored axis of scaling. To realize this, we propose LLM-as-a-Verifier, a general-purpose framework that delivers fine-grained feedback for agentic tasks. Unlike standard LM judges that output a single discrete score, our approach computes a continuous reward by taking the expectation over the distribution of scoring-token logits and enables verification scaling across multiple dimensions, including (1) score granularity, (2) repeated evaluation, and (3) criteria decomposition. When used as a trajectory reward model for test-time scaling, it achieves state-of-the-art performance on Terminal-Bench V2, SWE-Bench Verified, RoboRewardBench, and MedAgentBench. Beyond ranking, the fine-grained verifier signal can be used as a progress estimator, opening a path toward safer real-world deployment of autonomous agents. Finally, we demonstrate that LLM-as-a-Verifier can be used as a dense reward signal for RL, improving the sample efficiency of SAC and GRPO on robotics and mathematical reasoning benchmarks.

\clearpage

\section{Related Work}
\label{app:related-work}

\paragraph{Test-Time Scaling.}
Test-time scaling improves model performance by spending additional inference compute on deliberation, search, or candidate generation. One line of work improves a single response by eliciting intermediate reasoning with chain-of-thought~\citep{wei_chain--thought_nodate,kojima2022large}, decomposing problems into simpler subproblems~\citep{zhou2022least}, or marginalizing over sampled reasoning paths~\citep{wang_self-consistency_2023}. Another line searches over intermediate thoughts~\citep{yao2023tree,besta2023graph}, actions~\citep{yao_react_2023,zhou2023language}, or latent world states~\citep{hao2023rap} based on test-time feedback~\citep{weng2022selfverification,madaan_self-refine_nodate,shinn_reflexion_nodate,gou2023critic,agrawal_gepa_2025,novikov_alphaevolve_2025}. Repeated sampling and best-of-$N$ selection further scale candidate pools for code generation~\citep{li2022alphacode}, general reasoning~\citep{brown2024largelanguagemonkeysscaling,snell_scaling_2024}, parallel self-verification~\citep{singh_v_1_2026}, and inference-aware training~\citep{chow2024inferenceaware}. These approaches can expose substantial oracle headroom, but realizing this headroom requires a reliable selector. Our LLM-as-a-Verifier instead shows that verifier quality can be improved by scaling score granularity, repeated evaluation, and criteria decomposition, and that better verification directly improves best-of-$N$ selection for long-horizon decision making.

\paragraph{LLM-as-a-Judge.}
LLM-as-a-judge methods provide a scalable alternative to human evaluation by prompting large models to score or compare generated outputs. Probability- and form-filling-based evaluators extract richer scoring signals from LLMs~\citep{fu2023gptscore,liu2023geval}, while benchmark-style evaluators use LLM preferences to evaluate instruction-following systems~\citep{zheng2023judging,dubois2024alpacaeval,li2024arenahard}. A complementary line makes evaluation more fine-grained through skill decompositions~\citep{ye2023flask}, customized rubrics and specialized open judges~\citep{kim2024prometheus,kim2024prometheus2,li2024autoj,hu2024themis}, and hierarchical criteria~\citep{liu2024hdeval}. Other work trains scalable judge models~\citep{wang2024pandalm,zhu2025judgelm} or combines multiple judges through debate and weak-verifier ensembling~\citep{chan2023chateval,saad2025shrinking}. Recent studies have also analyzed judge reliability, including fairness and position biases~\citep{wang2024large,zeng2024llmbar}, cognitive and self-enhancement biases~\citep{koo2023cobbler,liu2023narcissistic}, general judge benchmarks~\citep{tan2025judgebench,huang2025empirical}, as well as domain-specific judge evaluation~\citep{jiang2025codejudgebench}. Multimodal judge models extend this paradigm to image and vision-language evaluation~\citep{chen2024mllmjudge,lee2024prometheusvision,xiong2024llavacritic}. Our work builds on these works but differs in setting, objective, and scaling characterization. Rather than evaluating isolated natural-language responses, LLM-as-a-Verifier verifies long-horizon agent trajectories involving tool use, code execution, robotics, and medical decision-making. Moreover, we systematically study how verification quality scales with score granularity, repeated evaluations, and criteria decomposition. 

\paragraph{Verifiable Reward.}
Reward models convert candidate solutions, actions, or trajectories into scalar feedback for selection, monitoring, or policy optimization. In language reasoning, learned verifiers have been trained as outcome reward models for final-answer selection~\citep{cobbe_training_2021}, process reward models for step-level supervision~\citep{uesato2022solving,lightman_lets_2023}, and generative verifiers that cast reward modeling as next-token prediction~\citep{zhang2025generativeverifiersrewardmodeling}. 
In robotics, reward signals have been derived from value-implicit visual representations~\citep{ma2022vip}, language-image reward representations~\citep{ma2023liv}, pretrained vision-language models~\citep{sontakke2023roboclip,rocamonde2023vlmrm,ICLR2025_54854cf1}, VLM feedback or preferences~\citep{wang2024rlvlmf}, LLM-generated reward code~\citep{yu2023languagetorewards,xie2023text2reward,ma2023eureka}, token-probability progress~\citep{chen2026topreward}, and trained general-purpose robotic reward models based on large-scale trajectory or preference data~\citep{zhang2025rewind,chen2025sarm,lee2026roboreward,liang2026robometer}.
Recent work has also developed action-level verifiers for guided sampling~\citep{nakamoto2024steering,liu2024bidirectional,kwok2025robomonkey}, runtime monitoring~\citep{agia2024unpacking}, and multimodal alignment~\citep{kwok2026scalingverificationeffectivescaling}.
A complementary line makes verification more reliable by factorizing holistic judgments into smaller checks~\citep{min2023factscore,fabbri2021qafacteval,manakul2023selfcheckgpt,liu2024hdeval,liu2026worldactionverifierselfimproving,tseng2026sc3}. Orthogonal to these methods, our work studies how verifier quality scales with score granularity, repeated evaluation, and criteria decomposition across multiple domains in a general-purpose framework.

\section{Acknowledgments}

We thank the members of the UC Berkeley Sky Computing Lab, Stanford Scaling Intelligence Lab, IRIS Lab, and Autonomous Systems Lab for their constructive feedback and informative discussions. This work was supported by Google; Google DeepMind; Google Cloud; Stanford HAI; DARPA (HR00112520038, Fallingwater); NSF (24-554, AIMing); NASA ULI; Schmidt Sciences; and Lightspeed. We also acknowledge the support of IBM and Felicis as members of Stanford HAI's Industry Affiliates Program.

\bibliography{ref}

\newpage
\clearpage
\appendix
\section*{Appendix}

\section{Limitations and Future Work}
\label{app:limitations}

The current framework has several limitations that suggest directions for future work. First, it assumes access to scoring-token logits, which excludes several frontier models available only through restricted APIs; in Appendix~\ref{app:closed-api} we describe a simple two-stage workaround that recovers most of the gain by routing the reasoning of a closed model through an open verifier whose logits are accessible. Second, the proposed scaling axes are not exhaustive: criteria decomposition could be learned or dynamically generated per domain rather than hand-designed, and repeated evaluation could be replaced with an adaptive compute allocation strategy guided by the verifier's own uncertainty. Finally, while we already show that the verifier can serve as a dense reward for reinforcement learning, our experiments are limited to single-turn settings; extending it to multi-turn RL---where the verifier supplies per-step rewards over long-horizon agentic rollouts to shape credit assignment across many interdependent actions---is a promising direction for future work.

\section{Additional Results and Analyses}
\label{app:additional}

\subsection{Agent Harness Generalization on Terminal-Bench V2}
\label{app:harness}

To verify that the gains of LLM-as-a-Verifier are not tied to a specific agent scaffold, we repeat the Terminal-Bench~V2 evaluation under two additional harnesses beyond the Capy scaffold used in our main results, each paired with the model its authors tuned for: Terminus-Kira with Claude Opus~4.6 and Terminus-2 with GPT-5.3-Codex. In every case we sample $N{=}5$ trajectories per task and apply the same Gemini~2.5~Flash verifier with $G{=}20$, $K{=}8$, and the three-criterion decomposition of Section~\ref{sec:criteria}; only the proposal generator and the harness change. Table~\ref{tab:harness-generalization} reports the resulting verifier accuracies alongside the official single-trajectory accuracies of Claude Opus~4.6 and Gemini~3.1 Pro under each harness.

\begin{table}[h]
\centering
\caption{\textbf{Harness generalization on Terminal-Bench~V2.} LLM-as-a-Verifier significantly boosts the accuracy on two additional harnesses, Terminus-2 ($71.2\%$) and Terminus-Kira ($79.4\%$).}
\label{tab:harness-generalization}
\small
\setlength{\tabcolsep}{6pt}
\begin{tabular}{lccc}
\toprule
Agent Harness & \textbf{LLM-as-a-Verifier} & Claude Opus~4.6 & Gemini~3.1 Pro \\
\midrule
Terminus-Kira (Opus~4.6)    & \textbf{79.4\%} & 74.7\% & 74.8\% \\
Terminus-2 (GPT-5.3-Codex)  & \textbf{71.2\%} & 62.9\% & 68.5\% \\
\bottomrule
\end{tabular}
\end{table}

The verifier delivers the same qualitative gain across both harnesses despite their setups, observation formats, and models. Terminus-Kira (Opus~4.6 proposals) gains $\sim\!5$ points over the strongest baseline, and Terminus-2 (GPT-5.3-Codex proposals)---the weaker harness in absolute terms---still gains $2.7$ points over Gemini~3.1 Pro and $8.3$ points over Claude Opus~4.6. The transfer indicates that the verifier reasons about \emph{terminal state and task progress} rather than about scaffold-specific syntactic patterns: the same prompt template generalizes across stylistically different rollouts.

\subsection{Probabilistic Pivot Tournament: Budget--Accuracy Trade-off}
\label{app:ppt}

We provide the full pseudocode for the LLM-as-a-Verifier pipeline. Algorithm~\ref{alg:testtime} embeds the fine-grained reward of Eq.~\ref{eq:main}---the expectation over the verifier's scoring-token distribution, averaged across the $C$ criteria and $K$ repeated evaluations---inside Probabilistic Pivot Tournament with ring-based pivot selection: a random Hamiltonian cycle gives every candidate one ``A'' position and one ``B'' position to cancel the verifier's positional bias, the top-$k$ candidates by ring mean preference become the pivot set $\mathcal{P}$, and each remaining candidate is compared only against $\mathcal{P}$ using the Bradley–Terry preference of Eq.~\ref{eq:pref}. The trajectory with the highest count-normalized score is returned, reducing the budget from $\mathcal{O}(N^2)$ to $\mathcal{O}(Nk)$ while concentrating verifications on the candidates most likely to be correct.

\vspace{2.5em}

\begin{algorithm}[h]
\caption{\textbf{Probabilistic Pivot Tournament with Ring-based Pivot Selection.} A random Hamiltonian cycle is first scored to give every candidate one ``A'' position and one ``B'' position; the top-$k$ candidates by ring mean preference become the pivots, and each remaining candidate is compared only against the pivot set using the soft preference probability of Eq.~\ref{eq:pref}.}
\label{alg:testtime}
\small
\begin{algorithmic}[1]
\Require Task $x$; generation policy $\pi_\theta$; verifier LM $p_\theta$;
score tokens $V_{\text{score}}$ with map $\phi$; criteria $\mathcal{C}$;
candidates $N$; repetitions $K$; pivots $k$
\Ensure Selected trajectory $\tau^\star$
\For{$i = 1, \ldots, N$} \Comment{candidate generation}
    \State Sample $\tau_i \sim \pi_\theta(\cdot \mid x)$
\EndFor
\State Initialize $w_i \gets 0,\ c_i \gets 0$ for $i\in\{1,\ldots,N\}$
\State Sample random permutation $\gamma$ of $\{1,\ldots,N\}$ \Comment{ring pass}
\State $\mathcal{E}_{\text{ring}} \gets \{(\gamma_t,\gamma_{t+1\,\mathrm{mod}\,N}) : t=1,\ldots,N\}$
\For{each pair $(i,j)\in\mathcal{E}_{\text{ring}}$}
    \State $(R_i, R_j) \gets \big(R(x,\tau_i),\, R(x,\tau_j)\big)$ \Comment{reward via Eq.~\ref{eq:main}}
    \State $p \gets \sigma(R_i - R_j)$ \Comment{Eq.~\ref{eq:pref}}
    \State $w_i \mathrel{+}= p,\ w_j \mathrel{+}= 1-p,\ c_i \mathrel{+}= 1,\ c_j \mathrel{+}= 1$
\EndFor
\State $\mathcal{P} \gets$ top-$k$ candidates by $w_i / c_i$ \Comment{ring-based pivot selection}
\State $\mathcal{E}_{\text{piv}} \gets \big(\{(i,p): i\!\notin\!\mathcal{P},\, p\!\in\!\mathcal{P}\}\cup\{(p_1,p_2)\!\in\!\mathcal{P}^2 : p_1<p_2\}\big)\setminus \mathcal{E}_{\text{ring}}$
\For{each pair $(i,j)\in\mathcal{E}_{\text{piv}}$} \Comment{$\mathcal{O}(Nk)$ pivot rounds}
    \State $(R_i, R_j) \gets \big(R(x,\tau_i),\, R(x,\tau_j)\big)$ \Comment{reward via Eq.~\ref{eq:main}}
    \State $p \gets \sigma(R_i - R_j)$ \Comment{Eq.~\ref{eq:pref}}
    \State $w_i \mathrel{+}= p,\ w_j \mathrel{+}= 1-p,\ c_i \mathrel{+}= 1,\ c_j \mathrel{+}= 1$
\EndFor
\State \Return $\tau^\star \gets \tau_{i^\star}$ where $i^\star \in \arg\max_i\, w_i / c_i$
\end{algorithmic}
\end{algorithm}

\vspace{2.5em}

We characterize the budget--accuracy trade-off of Probabilistic Pivot Tournament (PPT, Algorithm~\ref{alg:testtime}) and compare it against the V1 baseline~\citet{singh_v_1_2026}. We curate $N=20$ candidate trajectories per task on Terminal-Bench~V2 (89 tasks, Terminus-2 harness) and report the total number of queried pairs and selection accuracy in Table~\ref{tab:pivot-budget}. PPT improves steadily as the number of pivots increases, outperforming the V1 under comparable verification budgets. With only $k{=}3$, PPT already surpasses the best V1 result, achieving $66.17\%$ accuracy with $4{,}723$ queried pairs. Increasing the number of pivots further improves accuracy: $k{=}5$ reaches $66.27\%$ accuracy using $6{,}609$ pairs, while $k{=}9$ achieves $67.13\%$ accuracy with $9{,}630$ pairs, approaching full round-robin performance while using substantially fewer comparisons.

\clearpage

\begin{table}[h]
    \centering
    \caption{Probabilistic Pivot Tournament (PPT) scales with the number of pivots, achieving higher accuracy as $k$ increases while maintaining a low verification budget on Terminal-Bench~V2.}
    \label{tab:pivot-budget}
    \small
\begin{tabular}{lcc}
  \toprule
  Method & Pairs queried & Accuracy (\%) \\
  \midrule
  pass@1   & --- & $52.64$ \\
  \midrule
  V1 ($1N$ budget) \citep{singh_v_1_2026}   & $1{,}400$ & $64.64$ \\
  V1 ($3N$ budget) \citep{singh_v_1_2026}   & $4{,}200$ & $65.62$ \\
  V1 ($5N$ budget) \citep{singh_v_1_2026}   & $7{,}000$ & $65.85$ \\
  V1 ($7N$ budget) \citep{singh_v_1_2026}   & $9{,}800$ & $65.53$ \\
\midrule
    PPT \,$k{=}1$        & $2{,}570$ & $65.83$ \\
    PPT \,$k{=}3$        & $4{,}723$ & $66.17$ \\
    \textbf{PPT \,$k{=}5$ (ours)}   & $6{,}609$ & $\mathbf{66.27}$ \\
    \textbf{PPT \,$k{=}7$ (ours)}   & $8{,}242$ & $\mathbf{66.67}$ \\
    \textbf{PPT \,$k{=}9$ (ours)}   & $9{,}630$ & $\mathbf{67.13}$ \\                                           
    \midrule
  Full Round-Robin & $13{,}111$ & $67.42$ \\
  \bottomrule
  \end{tabular}
\end{table}

\subsection{LLM-as-a-Verifier as a Process and Outcome Reward Model}
\label{app:prm-orm}

We evaluate LLM-as-a-Verifier as a process reward model (PRM) for per-step verification on agentic tasks and as an outcome reward model (ORM) for Best-of-$N$ on coding and mathematics benchmarks. Used as a PRM, pass@1 scales monotonically with the number of sampled actions per step $k$: from $48.7\%$ to $55.7\%$ on TauBench and from $49.8\%$ to $54.3\%$ on Terminal-Bench as $k$ grows from $1$ to $9$ (Table~\ref{tab:prm_results}), at $3\times$ less compute than $V_1$~\citep{singh_v_1_2026}. Used as an ORM (Table~\ref{tab:orm_results}), it improves pass@1 by $9.5\%$ on SWE-Bench Lite, $18.5\%$ on AIME, and $21.3\%$ on HMMT over the base model and outperforms pointwise and pairwise baselines, while improving verification accuracy by $+6.2$ points over $V_1$ at $1.5\times$ less compute.

\begin{table}[!ht]
    \centering
    \caption{\textbf{LLM-as-a-Verifier as a PRM: pass@1 scales with the number of sampled actions per step.} Per-step verification with LLM-as-a-Verifier increases pass@1 monotonically as the number of candidate actions $k$ grows.}
    \label{tab:prm_results}
    \small
    \begin{tabular}{lcccc}
    \toprule
    Sampled actions per step & $k{=}1$ & $k{=}3$ & $k{=}5$ & $k{=}9$ \\
    \midrule
    TauBench (Gemini 2.5 Flash, pass@1)        & $48.7$ & $54.0$ & $55.1$ & $\mathbf{55.7}$ \\
    Terminal-Bench (Gemini 3 Flash, pass@1)    & $49.8$ & $51.4$ & $52.0$ & $\mathbf{54.3}$ \\
    \bottomrule
    \end{tabular}
\end{table}

\begin{table}[!ht]
    \centering
    \caption{\textbf{LLM-as-a-Verifier as an ORM improves pass@1 accuracy across coding and mathematics benchmarks.} Under Best-of-$N$ sampling, LLM-as-a-Verifier outperforms the base model and the pointwise/pairwise verifier baselines, with absolute improvements of $9.5\%$ on SWE-Bench Lite, $18.5\%$ on AIME, and $21.3\%$ on HMMT over the base model, while using a smaller verification budget than $V_1$~\citep{singh_v_1_2026}.}
    \label{tab:orm_results}
    \small
    \begin{tabular}{lccc}
    \toprule
    Method & SWE-Bench Lite & AIME & HMMT \\
    \midrule
    Base model                                                & $23.5$ & $71.5$ & $52.0$ \\
    Pointwise (LM judge, $N$ budget)                          & $29.7$ & $83.3$ & $63.5$ \\
    Pairwise ($V_1$~\citep{singh_v_1_2026}, $3N$ budget)      & $31.0$ & $86.0$ & $73.3$ \\
    \textbf{LLM-as-a-Verifier (ours, $2N$ budget)}            & $\mathbf{33.0}$ & $\mathbf{90.0}$ & $\mathbf{73.3}$ \\
    \bottomrule
    \end{tabular}
\end{table}

\subsection{Case Study: \texttt{query-optimize}}
\label{app:query-optimize-example}

This appendix expands the \texttt{query-optimize} case study from Section~\ref{sec:granularity} with the full task specification, ground-truth breakdown, and verifier reasoning trace. The trajectory pair is drawn under the OpenHands harness, with Claude~Opus~4.5 as the proposal generator and Gemini~2.5~Flash as the verifier.

\paragraph{Task instruction.}
\begin{quote}\small
You are given the Open English Wordnet (OEWN) database in SQLite format, located at \texttt{/app/oewn.sqlite}.

I implemented a sql query but it is not optimized. I have saved it in \texttt{/app/my-sql-query.sql}. Please make the query as efficient as possible while ensuring that the same output is produced.

Please save your solution in the file \texttt{/app/sol.sql}. This file must contain no comments, just one single sql query terminated by a semicolon.

Finally, please use sqlite syntax! Your code will not execute in sqlite if you use other dialects.
\end{quote}

\paragraph{Ground-truth breakdown.} Both candidate trajectories save an optimized SQL query and report a passing internal \texttt{diff} check, but Terminal-Bench's hidden grader assigns reward $1$ to one and reward $0$ to the other. The failing trajectory's verification step is methodologically unsound:
\begin{itemize}
    \item The agent attempts to run the original query against \texttt{/app/oewn.sqlite} twice ($60$s timeout, then $5$m$02$s) and aborts both times.
    \item To obtain a reference output, the agent runs \texttt{cp /app/oewn.sqlite /tmp/oewn\_test.sqlite} and then issues \texttt{CREATE INDEX} on \texttt{senses(wordid)}, \texttt{senses(synsetid)}, and \texttt{synsets(synsetid)} on the copy.
    \item Subsequent steps compare \emph{(original query on indexed copy)} to \emph{(optimized query on canonical, unindexed database)}---two different physical access paths whose tied \texttt{ORDER BY} keys are not guaranteed to break the same way at the \texttt{LIMIT 500} boundary.
    \item All verification artifacts are then deleted (\texttt{rm /tmp/oewn\_test.sqlite /tmp/original\_output.txt /tmp/optimized\_output.txt}), removing any evidence that could be re-checked.
    \item The optimized query is therefore never actually validated against the original on the canonical, unindexed database that Terminal-Bench's grader uses, and \texttt{sol.sql} fails the hidden test.
\end{itemize}
The correct trajectory simply waits the full $5$m$03$s for the original query to complete on the canonical database and runs a direct \texttt{diff} (exit code $0$) before exiting.

\paragraph{Gemini~2.5~Flash reasoning trace.} Across $16$ reasoning traces with thinking enabled, the verifier reliably identifies the soundness issue. Excerpt:
\begin{quote}\small\itshape
``The agent was unable to execute the original query to completion on the provided \texttt{/app/oewn.sqlite} database within a reasonable timeframe (it was interrupted twice, once after 60s and once after 5m2s). To obtain a reference output for comparison, the agent \emph{copied the database} (\ldots) and then \emph{added indexes} to this copied database. \ldots By modifying the database to obtain the reference output, the agent violated the implicit constraint of the task. Therefore, it did not correctly verify that its optimized query produces the same output as the original query running on the original, unindexed database.''
\end{quote}

\subsection{Scaling Repeated Evaluation and Visual Context on RoboRewardBench}
\label{app:roboreward}

\begin{figure}[h]
    \centering
    \includegraphics[width=\linewidth]{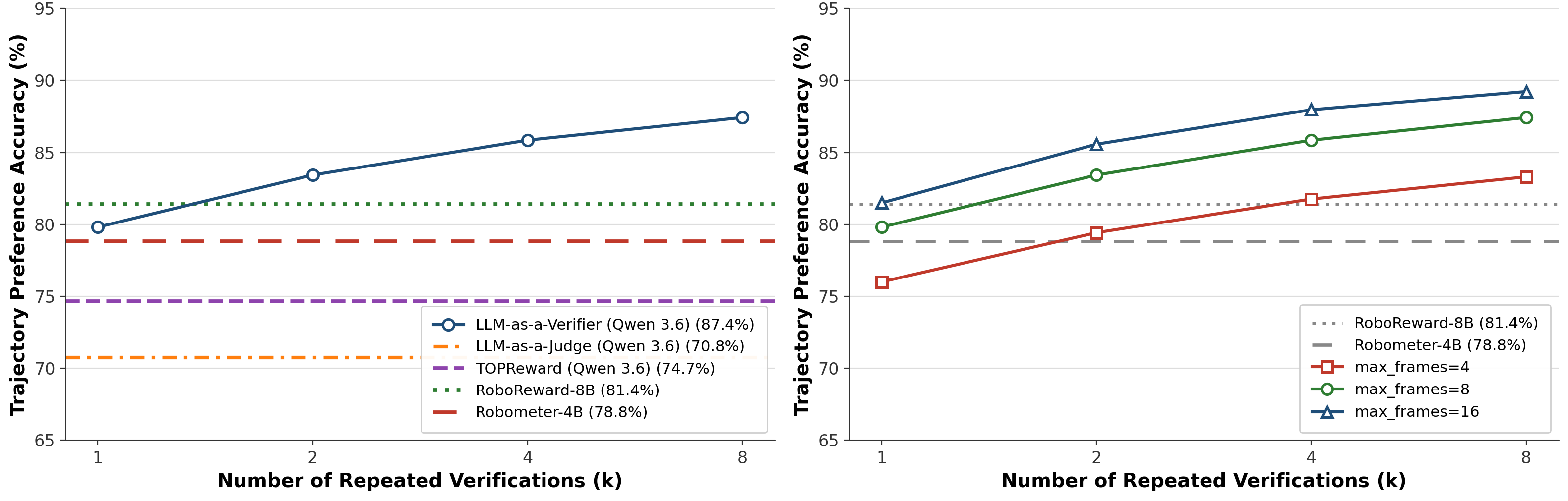}
    \caption{Trajectory-preference accuracy improves consistently as the number of repeated evaluations $k$ increases, rising from $81.5\%$ at $k{=}1$ to $87.4\%$ at $k{=}8$. LLM-as-a-Verifier outperforms LLM-as-a-Judge and prior reward models (TOPReward, RoboReward-8B, Robometer-4B) across all budgets, with gains saturating at larger $k$.}
    \label{fig:roboreward_repeated}
\end{figure}

Figure~\ref{fig:roboreward_repeated} extends the repeated-evaluation analysis of Section~\ref{sec:repeated} to robotic manipulation. Trajectory-preference accuracy on RoboRewardBench rises from $81.5\%$ at $K{=}1$ to $87.4\%$ at $K{=}8$ and saturates at large $K$ as the noise floor is reached. LLM-as-a-Verifier outperforms LLM-as-a-Judge and the trained baselines TOPReward, RoboReward-8B, and Robometer-4B at every budget, indicating that the gains of repeated evaluation transfer cleanly across modalities despite the change of input from text to multi-frame video.

\subsection{Recovering Continuous Rewards for Logit-Restricted Frontier Models}
\label{app:closed-api}

LLM-as-a-Verifier requires access to the verifier's scoring-token logits in order to evaluate Eq.~\ref{eq:main}. A growing class of frontier models, including GPT-5.5 and Claude Opus 4.7 exposes only sampled completions through their public APIs and does not return token-level logprobs, which precludes a direct in-place substitution of the verifier backbone. We describe a simple two-stage workaround that recovers most of the calibrated reward signal by decoupling reasoning from scoring.

\paragraph{Two-stage pipeline.} For each pair $(\tau_i, \tau_j)$, we first prompt the closed model (GPT-5.5) with our standard pairwise template and require it to emit a free-form $<$reasoning$>$\ldots$<$/reasoning$>$ block analyzing both trajectories before producing a discrete $1$--$10$ score. We then forward the task, both trajectories, and the closed-model reasoning to an open verifier (Gemini~2.5~Flash, $G{=}20$) and read its logprobs at the $<$score\_A$>$ and $<$score\_B$>$ positions to compute the continuous reward of Eq.~\ref{eq:main}. Stage~1 contributes domain-specific reasoning quality from the frontier model; stage~2 supplies the calibrated probability distribution that the closed API withholds.

\paragraph{Setup and Findings.} We evaluate on the same Terminal-Bench~V2 swing-pair suite used throughout Section~\ref{sec:repeated}. For each pair we generate $16$ independent reasoning traces with GPT-5.5 and $16$ independent stage-2 evaluations with Gemini~2.5~Flash, then average the $K\!\in\!\{1,2,4,8,16\}$ subsampled scores per trajectory and check whether $\bar R(x,\tau_{\text{correct}}) > \bar R(x,\tau_{\text{incorrect}})$. Mean accuracy and tie rate are estimated over $400$ bootstrap subsamples per $K$. Table~\ref{tab:closed-api} shows that the workaround dominates the discrete baseline at every budget. At $K{=}1$, routing the reasoning through Gemini~2.5~Flash recovers a $+5.2$-point accuracy gain over directly using GPT-5.5's integer scores ($80.1\%$ vs.\ $74.9\%$) and eliminates the $10.9\%$ tie rate that the closed model's coarse outputs incur. The continuous variant saturates almost immediately---accuracy moves only $1.1$ points from $K{=}1$ to $K{=}16$---whereas the discrete variant relies on heavy ensembling ($+4.2$ points from $K{=}1$ to $K{=}16$) primarily to break ties. Even at $K{=}16$, the continuous workaround leads by $+2.1$ points and retains zero ties.

\begin{table}[h]
    \centering
    \caption{ Accuracy and tie rate on Terminal-Bench~V2 as the number of repeated evaluations $K$ scales from $1$ to $16$. \textit{GPT-5.5 (Discrete)} averages the integer $1$--$10$ scores returned by GPT-5.5; \textit{GPT-5.5 $\rightarrow$ Gemini~2.5~Flash (Continuous)} forwards the GPT-5.5 reasoning to Gemini~2.5~Flash and reads continuous rewards from its scoring-token logits.}
    \label{tab:closed-api}
    \small
    \begin{tabular}{lcccc}
    \toprule
    & \multicolumn{2}{c}{GPT-5.5 (Discrete)} & \multicolumn{2}{c}{GPT-5.5 $\rightarrow$ Gemini~2.5~Flash (Continuous)} \\
    \cmidrule(lr){2-3} \cmidrule(lr){4-5}
    $K$ & Accuracy (\%) & Tie rate (\%) & Accuracy (\%) & Tie rate (\%) \\
    \midrule
    $1$  & $74.9$ & $10.9$ & $\mathbf{80.1}$ & $\mathbf{0.0}$ \\
    $2$  & $76.3$ & $\phantom{0}9.1$ & $\mathbf{80.5}$ & $\mathbf{0.0}$ \\
    $4$  & $77.6$ & $\phantom{0}7.0$ & $\mathbf{81.0}$ & $\mathbf{0.0}$ \\
    $8$  & $78.4$ & $\phantom{0}5.8$ & $\mathbf{80.9}$ & $\mathbf{0.0}$ \\
    $16$ & $79.1$ & $\phantom{0}5.0$ & $\mathbf{81.2}$ & $\mathbf{0.0}$ \\
    \bottomrule
    \end{tabular}
\end{table}
\subsection{LLM-as-a-Verifier as a Dense Reward for RL}
\label{app:rl}

This appendix details the RL experiments of Section~\ref{sec:rl}. In both settings the only difference between the baseline and our method is the reward: the policy, optimizer, hyperparameters, evaluation protocol, and random seeds are held fixed.

\paragraph{Off-policy RL (DSRL-SAC).} We fine-tune a $\pi_0$ policy on the LIBERO-90 \texttt{ketchup} task with DSRL-SAC. The verifier is Qwen 3.6 35B served over SGLang and queried with the task instruction and a uniform sub-sample of $N_f{=}10$ rendered frames per rollout; each frame's completion is scored on the granularity-$20$ scale of Eq.~\ref{eq:main}, decoded as the expectation over the top-$20$ scoring-token logprobs, and the evaluation is repeated $K=3$ times with coefficient $\lambda{=}1$ (Eq.~\ref{eq:rl-sac}). Reward shaping is applied to successful rollouts by default; relabeling all rollouts (success and failure) is an optional variant in the code base. Relabeled transitions are written to the SAC replay buffer and the critic is trained on the shaped returns. We run $5$ seeds per condition to $1.5\mathrm{M}$ environment steps and report the cross-seed mean of the simulator success rate.

\paragraph{On-policy RL (GRPO).} We fine-tune Qwen3-8B on Hendrycks MATH with GRPO using Tinker, with a group size of $M{=}16$, $64$ groups per optimization batch, learning rate $2\times10^{-5}$, and a maximum generation length of $512$ tokens. For each group of completions, Gemini~2.5~Flash scores the reasoning traces through the probabilistic pivot tournament (Eq.~\ref{eq:pref}). This preference is standardized within the group and added to the correctness and format reward with weight $\beta=0.1$ (Eq.~\ref{eq:rl-grpo}).

\end{document}